\pdfoutput=1

\documentclass[11pt]{article}

\usepackage[]{acl}

\usepackage{kotex}
\usepackage{bbding}

\usepackage{times}
\usepackage{latexsym}
\usepackage{amsmath,amssymb}
 
\usepackage[T1]{fontenc}

\usepackage[utf8]{inputenc}

\usepackage{microtype}
\usepackage{graphicx}
\usepackage{subcaption}
\usepackage{booktabs}

\usepackage{float}
\usepackage[export]{adjustbox}

\usepackage{tabularx}
\usepackage{arydshln}

\usepackage{makecell}

\usepackage{multicol}

\title{Few-shot Adaptation Works with UnpredicTable Data}

\author{Jun Shern Chan$^{1~2}$ ~~ Michael Pieler$^{1~2}$ ~~Jonathan Jao$^{1~2}$ ~~  \textbf{Jérémy Scheurer}$^{1~2}$ \\ \textbf{Ethan Perez}$^{1~2~3}$\thanks{~~Work done primarily at NYU and FAR.} \\
$^1$New York University, $^2$Fund for Alignment Research, $^3$Anthropic\\
  {\tt \{junshern,perez\}@nyu.edu} \\}

\begin{document}
\maketitle
\begin{abstract}

Prior work on language models (LMs) shows that training on a large number of diverse tasks improves few-shot learning (FSL) performance on new tasks. We take this to the extreme, automatically extracting 413,299 tasks from internet tables - orders of magnitude more than the next-largest public datasets. Finetuning on the resulting dataset leads to improved FSL performance on Natural Language Processing (NLP) tasks, but not proportionally to dataset scale. 
In fact, we find that narrow subsets of our dataset sometimes outperform more diverse datasets. For example, 
finetuning on software documentation from \texttt{support.google.com} raises FSL performance by a mean of +7.5\% on 52 downstream tasks, which beats training on 40 human-curated NLP datasets (+6.7\%).
Finetuning on various narrow datasets leads to similar broad improvements across test tasks, suggesting that the gains are not from domain adaptation but adapting to FSL in general.
We do not observe clear patterns between the datasets that lead to FSL gains, leaving open questions about why certain data helps with FSL.

\end{abstract}

\section{Introduction}

\begin{figure}[ht!]
    \centering
    \includegraphics[width=.5\textwidth,keepaspectratio]{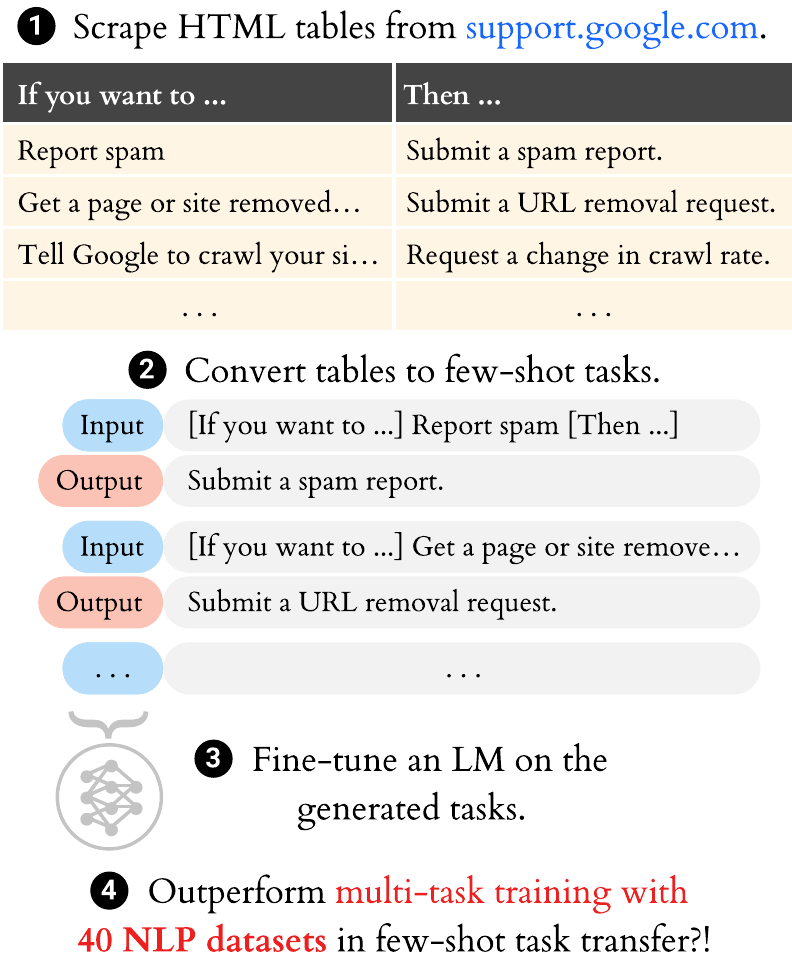}
    \caption{We convert a wide variety of tables into tasks for few-shot learning (FSL), then use these tasks via finetuning to adapt language models for FSL. Unexpected tables lead to strong task transfer results: finetuning GPT2 on software documentation from \texttt{support.google.com} outperforms finetuning on 40 curated NLP datasets on average across 52 test tasks, with strong improvements across diverse tasks including article classification (+47\%), sentiment classification (+31\%) and scientific question-answering (+23\%).}
    \label{fig:banner_diagram}
\end{figure}

\citet{brown2020language} showed that language models (LMs) learn to perform new tasks from a few examples (``few-shot learning''; FSL). Explicitly training LMs for FSL further improves performance \cite{min2021metaicl, chen2021meta}, and prior work has found that increasing the size and diversity of training tasks improves generalization to new tasks \cite{sanh2021multitask, aribandi2021ext5, aghajanyan2021muppet, wang2022benchmarking}. We push size and diversity to the extreme by finetuning on a large dataset of automatically-curated FSL tasks, and surprisingly find that certain narrow datasets of tasks (e.g. software documentation) outperform much larger and more diverse datasets.

Investigations into dataset size and diversity requires a large dataset of FSL tasks. To this end, we explore tables as a naturally-occurring source of diverse FSL tasks.
Given a table where each row is a list of fields, we hold out one row as the test example and treat all other rows as task training examples. We apply this idea to automatically convert internet tables into \texttt{UnpredicTable}\footnote{\href{https://github.com/JunShern/few-shot-adaptation}{https://github.com/JunShern/few-shot-adaptation}}, a dataset of 413,299 diverse few-shot tasks.
We finetune GPT-2 to perform a new task given a few task examples in its context~\citep[``MetaICL'';][]{min2021metaicl}.
Finetuning on \texttt{UnpredicTable} leads to strong FSL performance on average over 52 NLP test tasks, comparable to finetuning on human-curated NLP datasets.
However, the observed gains fall short of expectations for such a large dataset.

To understand why our gains were limited, we perform various ablations on dataset size, diversity, and content. In this process, we find that finetuning on narrow subsets of \texttt{UnpredicTable} outperforms finetuning on our diverse dataset and on curated NLP data.
Surprisingly, datasets that we handpick according to what we expect to be helpful are not strongly correlated with performance.
In fact, the training datasets that lead to strong improvements are often counterintuitive, covering trivia content (e.g. video games on \texttt{mmo-champion.com} and software documentation from \texttt{support.google.com}; see Fig. \ref{fig:banner_diagram}) that are unrelated to downstream test tasks.
Finetuning on these narrow datasets cause broad improvements similar to finetuning on curated NLP datasets when compared on the same test tasks. This suggests that these aren't domain- or task-specific improvements, but improvements in general few-shot ability (``few-shot adaptation'').
Our work calls into question common wisdom that adapting LMs to FSL requires diverse, high-quality training data.


\section{Web Tables as a Source of Few-Shot Learning Tasks}

We begin by describing FSL, which is the problem of learning from a small number of training examples. We make the case that web tables can be used as a diverse source of few-shot tasks. Then, we introduce our algorithm for converting tables into tasks and apply this to produce \texttt{UnpredicTable}, a dataset of 413,299 few-shot tasks.

\begin{figure*}[ht!]
    \centering
    \includegraphics[width=1\textwidth,keepaspectratio]{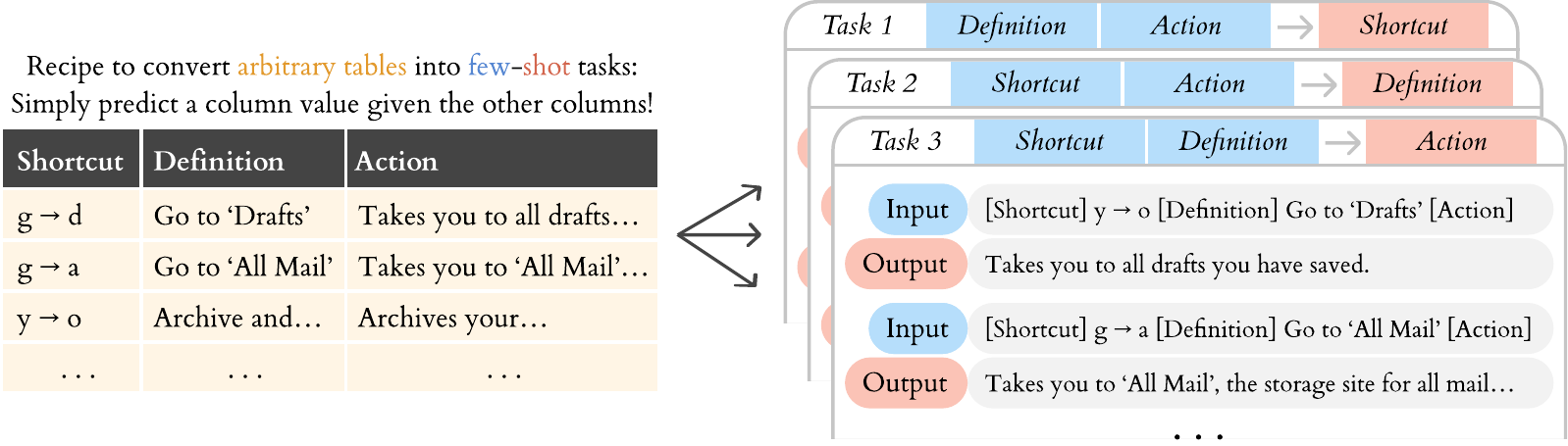}
    \caption{An algorithm to convert tables into tasks for FSL: Given the task of "Predict this column value given the other column values as input," each row in the table can be used as an example for that task.}
    \label{fig:tables-to-tasks}
\end{figure*}

\subsection{Few-Shot Learning Tasks}
\label{sec:few_shot_tasks}


We define a task $T$ as a set of input-output pairs $T = \{(x_i, y_i)\}^k_{i=1}$ where inputs $x_i$ map to outputs $y_i$. Task types can be very diverse, from question-answering (Questions $\rightarrow$ Answers), to summarization (Books $\rightarrow$ Summaries), to translation (French $\rightarrow$ English). In FSL, $k$ is small. LMs can be used to perform FSL by providing \(k\) known example pairs $\left\{(x_i,y_i): i=1, \ldots, k\right\}$ in the LM context at inference time. Then, we give the model a new example $x_\text{target}$ for which $y_\text{target}$ is unknown, and we use the model to predict $y_\text{target}$.

\subsection{Tables Dataset}
\label{sec:table_dataset}
Motivated by prior work on FSL adaptation \cite{min2021metaicl, chen2021meta} and multi-task learning \cite{sanh2021multitask, aribandi2021ext5, aghajanyan2021muppet}, we hypothesize that we can extend the results of multi-task FSL finetuning with an even larger set of few-shot tasks. We make the case that web tables are a large and diverse source of few-shot tasks. Consider a table where each row is an instance of a similar class and columns describe the attributes of an instance. We use each row as an example of a task, where the task is filling in missing attributes in a row. For a table with \(k\) rows, each table becomes a \(k\)-shot dataset for a particular task.

As a source of table data, we use tables from the English-language Relational Subset of the WDC Web Table Corpus 2015 (WTC)\footnote{\href{http://webdatacommons.org/webtables/2015/EnglishStatistics.html}{webdatacommons.org/webtables/2015/EnglishStatistics.html}}. The WTC dataset was extracted from the July 2015 Common Crawl web corpus, and contains 50M tables from 323K web domains. We focus on relational tables, which describe a set of similar items along with their attributes. For example, a table listing national dishes by country is a relational table. On the other hand, a table describing a single item where each row describes a different attribute is not relational. WTC also provides helpful metadata including the source URL, title, and header rows.

\subsection{Turning Tables Into Tasks}
\label{sec:converting_tables}
In practice, there are important design choices for converting a table into a task of input-output pairs. Here, we describe our chosen procedure. We start with the assumption that items in the relational table are listed row-wise (as in Fig. \ref{fig:tables-to-tasks}) instead of column-wise. Where necessary, we transpose the tables to suit our requirement. To convert a row into an input-output task pair, we consider a single column as a potential output target $y_i$ and concatenate the remaining columns to form the input $x_i$. For additional context, we prefix each value with its column header (see Fig. \ref{fig:tables-to-tasks}). Since any column is a potential output target, we create multiple tasks per table. For example, a table with 3 columns A, B, and C may be cast as three different tasks: $P(A | B, C)$, $P(B | A, C)$ and $P(C | A, B)$.

\paragraph{Filtering tables} We reject tables with fewer than 2 unique columns (one for the task output and at least one more for the input) or 6 unique rows (at least 5 examples + 1 target row). We find a large number of tables containing junk data or only numerical values. To remove these, we reject tables with $\ge 20\%$ of tokens tagged as either \textit{Numeral}, \textit{Proper Noun}, \textit{Symbol}, \textit{Punctuation}, or \textit{Other} by the \texttt{spaCy} part-of-speech classifier.\footnote{\href{https://spacy.io/usage/linguistic-features\#pos-tagging}{spacy.io/usage/linguistic-features\#pos-tagging}} The tables that pass this filtering stage are converted into tasks.

\paragraph{Filtering tasks} Given a set of candidate tasks, we require that the output space contains at least two unique answers, and reject tasks with severe class imbalance.\footnote{We measure class imbalance using \href{https://stats.stackexchange.com/questions/239973/a-general-measure-of-data-set-imbalance}{Shannon Diversity Index} and reject scores lower than 0.7.} To narrow our scope to tasks with a single correct answer, we reject tasks where any input appears more than once with different outputs. Finally, we only accept up to 2500 tasks per website to counter imbalance\footnote{Without data rebalancing, cappex.com makes up 41\% of the tasks.} in the source website of generated tasks. Appendix \ref{sec:tables_to_tasks_filtering} shows the breakdown of filtered tables and tasks at each stage.

We apply our tables-to-tasks procedure to produce \texttt{UnpredicTable}, a dataset with 413,299 tasks from 23,744 unique websites. The shape of our dataset is very different from most NLP datasets: NLP datasets typically contain a handful of tasks, with thousands of examples per task. On the other hand, \texttt{UnpredicTable} contains 400K tasks but most tasks have fewer than 50 examples. Thus, our dataset has a large variety of tasks but each task has limited training examples, true to the small-$k$ FSL setting. Our data-generation code and corresponding dataset are open-source.\footnote{\href{https://github.com/JunShern/few-shot-adaptation}{github.com/JunShern/few-shot-adaptation}}

\section{Multitask Training with Few-shot Tasks for Few-shot Adaptation}
\label{sec:finetuning}

The shape of our dataset makes it suitable for multitask learning algorithms. In multitask learning, we have a training dataset $\mathcal{D}_\text{train} = \{T_i\}^{M_\text{train}}_{i=1}$ containing $M_\text{train}$ training tasks $T$, and a test dataset $\mathcal{D}_\text{test}$ with $M_\text{test}$ tasks which are disjoint to $\mathcal{D}_\text{train}$. The key idea is to use $\mathcal{D}_\text{train}$ to train a model to be generalizable to new tasks in $\mathcal{D}_\text{test}$.

Here, we focus on the MetaICL algorithm \cite{min2021metaicl} for few-shot adaptation, which has shown strong FSL results across a variety of downstream tasks. We show additional experiments on the CrossFit \cite{ye2021crossfit} and FLEX \cite{bragg2021flex} benchmarks in Appendix \ref{sec:additiona_setups}, to study the generalization of our results across different models, training algorithms and test tasks.

\subsection{MetaICL}

MetaICL \cite{min2021metaicl} trains LMs to predict the output for a target input, given a few input-output pairs provided in the LM context. On each training iteration, one task $T_i$ is sampled from $\mathcal{D}_\text{train}$ and $k + 1$ training examples $\{(x_1, y_1), \dots,(x_{k+1}, y_{k+1})\}$ are sampled from $T_i$. MetaICL trains an LM with parameters $\theta$ to maximize $\log P(y_{k+1} | x_1, y_1, \dots, x_k, y_k, x_{k+1})$. At test time, for a new task in $\mathcal{D}_\text{test}$ we draw a set of examples $\{x_1, y_1, \dots, x_k, y_k\}$ and a query $x_{k+1}$. Given this context, the LM uses $\theta$ to select the most likely $y_{k+1}$ from a discrete set of possible labels.




\subsection{Experiments}

Here, we investigate how finetuning on \texttt{UnpredicTable} compares to finetuning on human-curated NLP datasets. We finetune the 774M parameter pretrained GPT2-large LM \cite{radford2019language}, following \citet{min2021metaicl}. See Appendix \ref{sec:metaicl_details} for details on our hyperparameter and finetuning setup.

\paragraph{NLP datasets and evaluation settings}
\citet{min2021metaicl} use 142 unique NLP tasks from \citet{ye2021crossfit} and \citet{khashabi2020unifiedqa} to form $\mathcal{D}_\text{train}$ and $\mathcal{D}_\text{test}$ for 5 different NLP task categories: 26 \textit{Low Resource} (LR) tasks with <1000 examples per task, 8 \textit{Natural Language Inference} (NLI) tasks to test entailment between a premise and hypothesis clause, 4 \textit{Paraphrase} (Para) tasks that test the equivalence of two differently-worded phrases, 20 \textit{Classification} (Class) tasks, and 22 \textit{Question-Answering} (QA) tasks. We show results on each category. See Appendix \ref{sec:metaicl_details} for a full list of tasks.

\paragraph{MetaICL methods}
MetaICL evaluates performance on each task category in two ways.
First, they consider an out of distribution (``OOD'') setting, where they finetune a model on a dataset $\mathcal{D}_\text{train}$ consisting of tasks from all other categories excluding the target task category.
Second, for \textit{Class} and \textit{QA} categories, they consider an in-domain (``IID'') setting, where they finetune a model on a dataset $\mathcal{D}_\text{train}$ consisting of only tasks from the same category as the target task category.


\paragraph{Our dataset} We sample $M=5000$ tasks from \texttt{UnpredicTable}, choosing $M$ based on results on a development set of tasks (Appendix \ref{sec:metaicl_details}). We refer to this dataset as \texttt{UnpredicTable-5k}. \citet{min2021metaicl} train one model per task category, while we fine-tune a single GPT2-large model on \texttt{UnpredicTable-5k} and test the resulting model on all task categories.

\subsection{Results}

\begin{table}[h]
\resizebox{\columnwidth}{!}{%
\begin{tabular}{lccccc|}
\cline{2-6}
\multicolumn{1}{l|}{} & \multicolumn{5}{c|}{\textbf{Task category {[}\# test tasks{]}}} \\ \hline
\multicolumn{1}{|l|}{\textbf{Method}} & \multicolumn{1}{c|}{\textbf{LR}} & \multicolumn{1}{c|}{\textbf{Class}} & \multicolumn{1}{c|}{\textbf{QA}} & \multicolumn{1}{c|}{\textbf{NLI}} & \textbf{Para} \\ \hline
\multicolumn{1}{|l|}{\textit{GPT2 0-shot}} & \multicolumn{1}{c|}{34.9} & \multicolumn{1}{c|}{34.2} & \multicolumn{1}{c|}{40.4} & \multicolumn{1}{c|}{25.5} & 34.2 \\ \hline
\multicolumn{1}{|l|}{\textit{GPT2 k-shot}} & \multicolumn{1}{c|}{38.2} & \multicolumn{1}{c|}{37.4} & \multicolumn{1}{c|}{40.2} & \multicolumn{1}{c|}{34} & 33.7 \\ \hline
\multicolumn{6}{|l|}{\textit{MetaICL k-shot trained with}} \\ \hline
\multicolumn{1}{|l|}{\textit{NLP (OOD)}} & \multicolumn{1}{c|}{43.2} & \multicolumn{1}{c|}{38.2} & \multicolumn{1}{c|}{38.7} & \multicolumn{1}{c|}{\textbf{49}} & 33.1 \\
\multicolumn{1}{|l|}{\textit{NLP (IID)}} & \multicolumn{1}{c|}{-} & \multicolumn{1}{c|}{43.4} & \multicolumn{1}{c|}{\textbf{45.9}} & \multicolumn{1}{c|}{-} & - \\ \hline
\multicolumn{1}{|l|}{\vtop{\hbox{\strut \textit{UnpredicTable-5k}}\hbox{\strut \textit{(our dataset)}}}} & \multicolumn{1}{c|}{\textbf{43.7}} & \multicolumn{1}{c|}{\textbf{46.1}} & \multicolumn{1}{c|}{42.3} & \multicolumn{1}{c|}{36.3} & \textbf{45.7} \\ \hline
\end{tabular}%
}
\caption{Columns represent different test settings; rows represent different methods. \textit{MetaICL k-shot} with finetuning on our dataset improves pretrained model performance (\textit{GPT2 k-shot}) on all test categories. Furthermore, finetuning on our tasks beats finetuning on out-category NLP datasets (\textit{OOD}) on 4/5 settings, and in-category NLP datasets (\textit{IID}) on 1/2 settings.}
\label{tab:metaicl_results}
\end{table}

For each task category, we compute the mean accuracy per task and report the average task accuracy for all tasks in the category. Tab. \ref{tab:metaicl_results} shows the results.
MetaICL finetuning on our table tasks improves FSL performance on all test settings. Furthermore, finetuning on our dataset outperforms finetuning on OOD NLP tasks on 4/5 settings, and IID NLP tasks on 1/2 settings.
Overall, finetuning on our data results in comparable performance to finetuning on curated NLP tasks.

\section{Why Is UnpredicTable Helpful?}
\label{sec:distributions}

\begin{figure*}[ht!]
  \includegraphics[width=\textwidth]{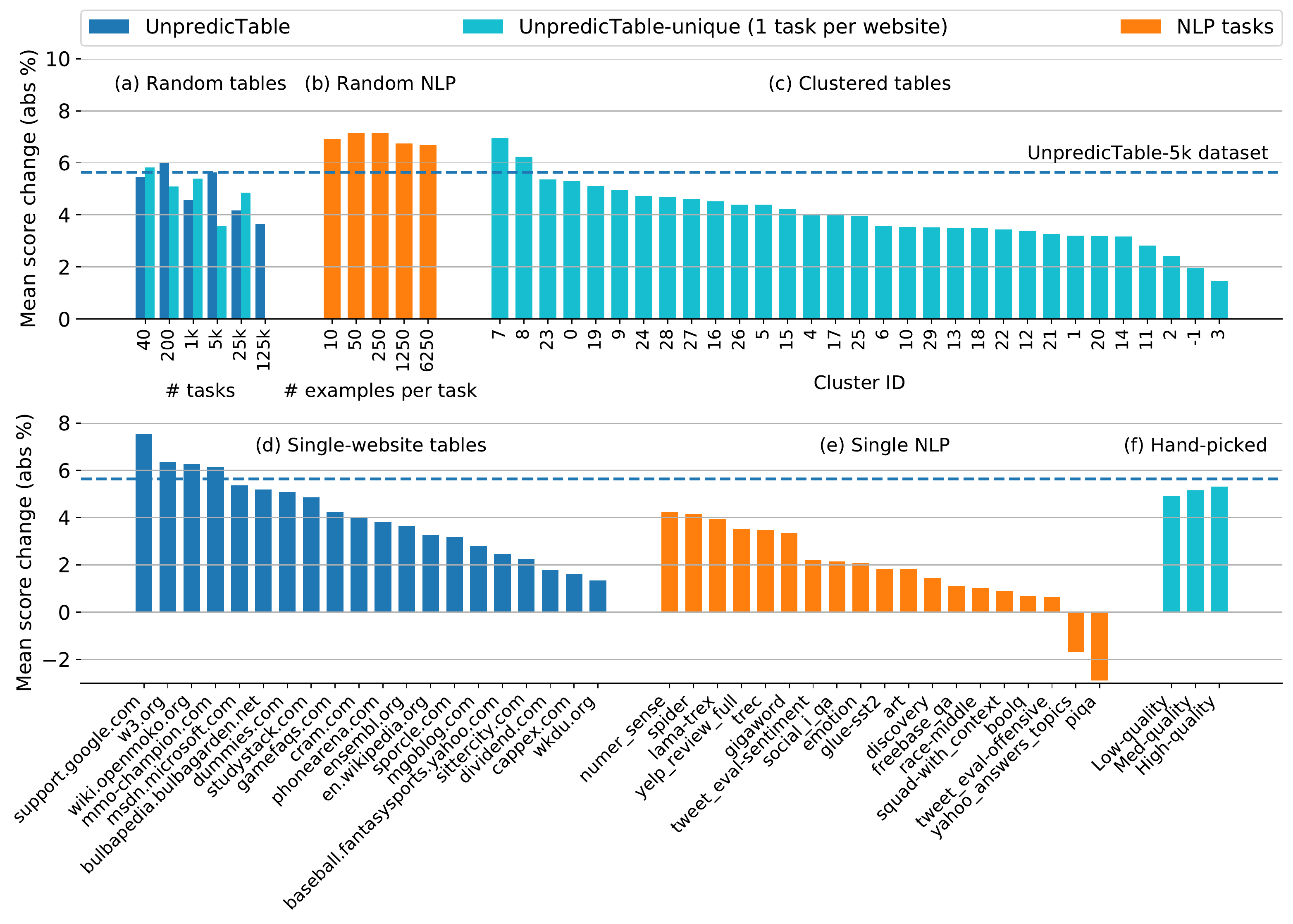}
  \caption{Each bar represents a GPT2 model finetuned on a different dataset. The y-axis shows mean improvement of a finetuned LM over the pretrained LM. \textbf{Comparing dataset helpfulness:} Datasets made of diverse tasks from UnpredicTable (a) and NLP datasets (b) lead to +5--7\% improvement. Narrow clusters (c) and websites (d) within UnpredicTable vary significantly, with the best narrow datasets matching the best multi-task NLP datasets (b).}

  \label{fig:task_distributions}
\end{figure*}

To understand why \texttt{UnpredicTable} is helpful training data, we construct subsets of the dataset varying features we wish to study.
For each sub-dataset, we finetune on that dataset individually following the setup as before (Appendix \ref{sec:metaicl_details}) and measure FSL performance on MetaICL test tasks from all categories (52 total). All experiments are repeated for 3 random seeds to minimize the effects of random task sampling in each dataset. We report the mean accuracy from each experiment in Fig. \ref{fig:task_distributions}. We discuss our results in the following sections.


\subsection{Does increasing dataset size improve finetuning performance?} 

Fig. \ref{fig:task_distributions}a shows FSL performance for differently-sized datasets randomly sampled from \texttt{UnpredicTable}. Each dataset has a maximum number of examples per task $N=10$ and varies the number of tasks $T$. 
Increasing the number of tasks from $T=40$ does not help and performance deteriorates beyond $T=5000$, contrary to results in \citet{wang2022benchmarking}.\footnote{For additional dataset scaling results, we randomly sample human-curated NLP tasks from the MetaICL training set (Fig. \ref{fig:task_distributions}b). Since there are only 90 NLP training tasks, we use $T=40$ tasks and vary $N$ to match the total number of examples in Fig. \ref{fig:task_distributions}a. At an equal number of tasks and examples per task ($T=40$, $N=10$), NLP datasets outperform our dataset by $\sim 1\%$. (The results in Tab. \ref{tab:metaicl_results} differ due to the choices of train and test tasks in different task categories.)}
Overall, the number of tasks does not seem to be the key factor for our finetuning transfer success.


\subsection{Does dataset diversity improve performance?}
Next, we study the effect of task diversity on FSL performance. Tasks from the same website tend to be similar in content, so we construct more diverse datasets by sampling tasks from \texttt{UnpredicTable-unique}, a version of \texttt{UnpredicTable} filtered to have a maximum of one task per website (vs. up to 2500 in \texttt{UnpredicTable}). 
Fig. \ref{fig:task_distributions}a shows that the difference between \texttt{UnpredicTable-unique} and \texttt{UnpredicTable} at matching sizes is small, suggesting that dataset diversity is not an important factor for our finetuning transfer success.

To examine narrow datasets in contrast to the uniformly-sampled ones, we consider 3 types of datasets grouped by content. We sample tasks from 20 websites of different genres, forming a dataset from each website (Fig. \ref{fig:task_distributions}d). Secondly, we also form datasets of semantically similar tasks by clustering \texttt{UnpredicTable-unique} tasks into 30 clusters using HDBSCAN\footnote{See Appendix \ref{sec:clustering_details} for details of our clustering setup.} \cite{mcinnes2017hdbscan} (Fig. \ref{fig:task_distributions}c).
Finally, we also sample 20 NLP tasks from the 90 MetaICL training tasks and use each task as a separate training dataset (Fig. \ref{fig:task_distributions}e). Single-website and single-NLP datasets have $T \times N = 10000$ total examples, and cluster datasets have different $T$ due to the clustering algorithm.

We find there is significant variance among the narrow datasets. Some single-website or cluster datasets are better than diverse datasets, such as \texttt{support.google.com} which is our best dataset overall (even outperforming diverse NLP datasets). This suggests that diverse task datasets are less important than careful selection of a narrow training dataset for FSL improvement.


\subsection{Can we select good tasks by hand?}
\citet{padmakumar2022exploring} found that some training tasks can negatively impact downstream performance, which could explain why aggregating many random tasks may be less successful than individual tasks.
We manually categorize 2,000 tasks from \texttt{UnpredicTable-unique} into High, Mid, and Low-quality.\footnote{See Appendix \ref{sec:task_quality_annotations} for details of our annotation setup.} We define low-quality tasks as tasks where the content is junk or relies on missing context. High-quality tasks are ones where an annotator could pick the correct answer from a list of options, and tests useful abilities (logic, general knowledge, comprehension, etc.). Mid-quality tasks are the remaining tasks. For each class, we randomly sample $T=200$ tasks to form its own dataset.

Surprisingly, our manual annotations of quality are not strongly correlated with downstream task performance (Fig. \ref{fig:task_distributions}f). Our handpicked dataset of high-quality tasks does not even surpass the scores of randomly-sampled tasks, and the difference in performance between our low and high-quality datasets are <1\%. These results suggest that tasks that look helpful are not necessarily helpful.


\subsection{How do helpful and unhelpful tasks look?}
\begin{centering}
\begin{table}[ht!]
\begin{tabular}[t]{|c|p{0.78\linewidth}|}
\hline
\multicolumn{2}{|c|}{\textbf{\textit{Examples of Helpful Tasks}}} \\
\hline
\multicolumn{2}{|l|}{\texttt{w3.org}} \\
\hline
input & \small{[Keyword] password [Data type] Text with no line breaks (sensitive information) [Control type] A text field that obscures data entry [State]} \\
output & \small{Password}  \\ [5pt]
\hline
\multicolumn{2}{|l|}{\texttt{bulbapedia.bulbagarden.net}} \\
\hline
input & \small{[Move] Odor Sleuth [Effect]} \\
output & \small{Never ends, screen freezes with the words "Wild/Foe (Pokémon) used Odor Sleuth!"} \\ [5pt]
\hline
\multicolumn{2}{|l|}{\texttt{cluster 7}} \\
\hline
input & \small{[Cookie] guest\_id, ki [Information]} \\
output & \small{These cookies allow you to access the Twitter feed on the homepage.}  \\ [5pt]
\hline
\multicolumn{2}{|c|}{\textbf{\textit{Examples of Unhelpful Tasks}}} \\
\hline
\multicolumn{2}{|l|}{\texttt{wkdu.org}} \\
\hline
input & \small{[Artist] Noah and the Whale [Title]} \\
output & \small{5 Years Time}  \\ [5pt]
\hline
\multicolumn{2}{|l|}{\texttt{cappex.com}} \\
\hline
input & \small{[Comments] The school is located near town so anything you would want to do is just an easy ten minute drive away. [Categories]}  \\
output & \small{What to do for fun}  \\ [5pt]
\hline
\multicolumn{2}{|l|}{\texttt{yahoo\_answers\_topics}} \\
\hline
input & \small{question\_title: bungee jumping site in victoria??? [SEP] question\_content: i am trying to find a site for bungee jumping \textit{{... (Truncated)}}} \\
output & \small{Sports}  \\ [5pt]
\hline
\end{tabular}
\caption{Helpful and unhelpful datasets are highly varied and do not always match our intuitions on task quality.}
\label{tab:task_examples}
\end{table}
\end{centering}


We look for features of helpful and unhelpful datasets with examples from cluster, single-website and single-NLP datasets.
4/5 of the most helpful datasets are software-related. \texttt{support.google.com}, \texttt{w3.org} and \texttt{wiki.openmoko.org} contain software documentation; \texttt{cluster 7} describes information related to internet cookies. Unhelpful datasets are more varied. The two least-helpful datasets are NLP datasets: \texttt{piqa} (question-answering task for physical knowledge) and \texttt{yahoo\_answers\_topics} (topic-classification task) both yield negative transfer results. The least helpful table datasets include highly-repetitive software tables (\texttt{cluster 2 \& 3}), tasks classified as noise by the clustering algorithm (\texttt{cluster -1}), college review posts (\texttt{cappex.com}), and music database entries (\texttt{wkdu.org}).

The top datasets appear unrelated to our test tasks (e.g. there are no software-related test tasks). Additional examples highlight this: \texttt{mmo-champion.com} and \texttt{bulbapedia.bulbagarden.net} are video game trivia sites that do not seem useful for other tasks, yet these datasets are on par with \texttt{UnpredicTable-5k}. Conversely, websites containing high-quality question-answer pairs such as \texttt{cram.com} and \texttt{studystack.com}, as well as \texttt{en.wikipedia.org} which contains many real-world facts, yield subpar improvements.
We include examples of helpful and unhelpful tasks in Tab. \ref{tab:task_examples}, and more examples in Appendix \ref{sec:task_examples}.


\subsection{Which tasks are our datasets helpful for?}

\begin{figure}[ht!]
    \centering
    \includegraphics[width=.48\textwidth]{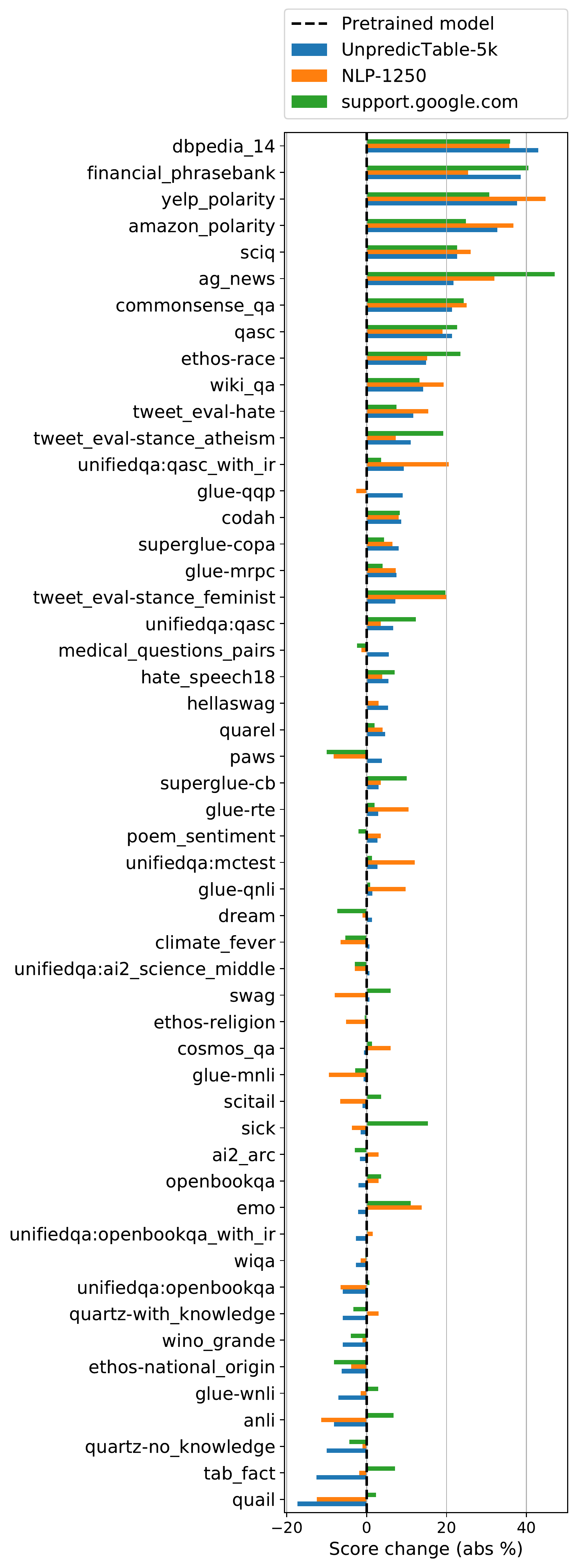}
    \caption{Breakdown of model scores across 52 test tasks for models finetuned on three different datasets. Scores are relative to the initial pretrained model.}
    \label{fig:breakdown}
\end{figure}

\begin{table}[ht]
\resizebox{\columnwidth}{!}{%
\begin{tabular}{l|rrr|}
\cline{2-4}
 & \multicolumn{1}{l|}{\textbf{Table-5k}} & \multicolumn{1}{l|}{\textbf{NLP-1250}} & \multicolumn{1}{l|}{\textbf{support.google}} \\ \cline{2-4} 
 & \multicolumn{3}{c|}{\textit{Test tasks counts (\# out of 52)}} \\ \hline
\multicolumn{1}{|l|}{\textbf{\textgreater Pretrained}} & \multicolumn{1}{r|}{33} & \multicolumn{1}{r|}{32} & \textbf{37} \\ \hline
\multicolumn{1}{|l|}{\textbf{\textless Pretrained}} & \multicolumn{1}{r|}{19} & \multicolumn{1}{r|}{20} & \textbf{15} \\ \hline
\multicolumn{1}{|l|}{\textbf{\begin{tabular}[c]{@{}l@{}}\textgreater Chance\\ (pre: 23)\end{tabular}}} & \multicolumn{1}{r|}{23} & \multicolumn{1}{r|}{31} & \textbf{34} \\ \hline
 & \multicolumn{3}{c|}{\textit{Score change (finetuned - pre) (\%)}} \\ \hline
\multicolumn{1}{|l|}{\textbf{Mean}} & \multicolumn{1}{r|}{+5.6} & \multicolumn{1}{r|}{+6.7} & \textbf{+7.5} \\ \hline
\multicolumn{1}{|l|}{\textbf{Median}} & \multicolumn{1}{r|}{+2.8} & \multicolumn{1}{r|}{+3.5} & \textbf{+3.6} \\ \hline
\multicolumn{1}{|l|}{\textbf{Max}} & \multicolumn{1}{r|}{+43.0} & \multicolumn{1}{r|}{+44.7} & \textbf{+47.1} \\ \hline
\multicolumn{1}{|l|}{\textbf{Min}} & \multicolumn{1}{r|}{-17.3} & \multicolumn{1}{r|}{-12.5} & \textbf{-10.0} \\ \hline
\end{tabular}%
}
\caption{\textit{Top (counts)}: First two rows indicate the number of test tasks that improved or not (vs the pretrained model) after finetuning. Third row shows the number of test tasks that score greater than random chance (on multiple-choice answers). Fine-tuning improves the pretrained model on more than 60\% of test tasks. \textit{Bottom (scores)}: Improvements are not evenly distributed; the maximum score increase on \texttt{support.google.com} is +47.1\% but median improvement is only +3.6\%.}
\label{tab:results_statistics}
\end{table}

\begin{figure*}[ht!]

\centering
\resizebox{\textwidth}{!}{%
\begin{tabular}{|c|c|c|c|c|c|c|}
\hline
\textbf{Task} & \textbf{Type} & \textbf{Output space} & \textbf{Chance (\%)} & \textbf{Median (\%)} & \textbf{Max (\%)} & \textbf{Best dataset} \\ \hline
ag\_news & News class & World / Sports / Business / SciTech & 25 & 42 (+29) & 63 (+50) & cluster 7 \\ \hline
dbpedia\_14 & Wikipedia class & 14 classes (plant / athlete / ...) & 7 & 31 (+25) & 47 (+42) & w3.org \\ \hline
commonsense\_qa & General QA & MCQ & 20 & 44 (+23) & 51 (+30) & cluster 12 \\ \hline
sciq & Scientific QA & MCQ & 25 & 81 (+23) & 87 (+29) & cluster 0 \\ \hline
amazon\_polarity & Review class & positive / negative & 50 & 77 (+18) & 92 (+34) & cluster 7 \\ \hline
qasc & General QA & MCQ & 13 & 30 (+17) & 38 (+25) & cluster 8 \\ \hline
financial\_phrasebank & Financial class & positive / negative / neutral & 33 & 41 (+14) & 68 (+40) & support.google.com \\ \hline
tweet\_eval-stance\_atheism & Tweet class & none / against / favor & 33 & 31 (+13) & 44 (+25) & msdn.microsoft.com \\ \hline
yelp\_polarity & Review class & positive / negative & 50 & 61 (+12) & 84 (+36) & w3.org \\ \hline
ethos-race & Hate speech class & true / false & 50 & 43 (+12) & 55 (+23) & support.google.com \\ \hline
\end{tabular}%
}
\caption{\textit{Most-improving tasks in the MetaICL test set:} The tasks span a wide variety of topics and output spaces. There is no clear connection to the training datasets that most strongly improve FSL performance (\textbf{Best dataset}), yet score improvements are significant. We show absolute scores for random \textbf{Chance} as well as the \textbf{Median} and \textbf{Max} scores across different training datasets. Improvements w.r.t. to the pretrained model are shown in parentheses.}
\label{tab:best_tasks}

\end{figure*}

Here, we investigate which test tasks benefit from our finetuning. Fig \ref{fig:breakdown} shows score improvements on all 52 test tasks relative to the pretrained model after finetuning on \texttt{UnpredicTable-5k}, \texttt{NLP-1250}\footnote{Random NLP tasks with $T=40, N=1250$ to match the total number of examples in \texttt{UnpredicTable-5k}.}, and \texttt{support.google.com}. Summary statistics are shown in Tab. \ref{tab:results_statistics}. Across the 3 datasets, 60-70\% of tasks have improved scores over the pretrained model.
The distribution of test score improvements appear to be highly concentrated on a few tasks, with 20\% of test tasks accounting for 60-80\% of all improvement. The median score change for \texttt{UnpredicTable-5k} is only +2.8\%, though the max is +43.0\%.

Fig. \ref{tab:best_tasks} shows the 10 most-improving test tasks (median improvement across all 90 training datasets in Fig. \ref{sec:distributions}). The tasks are highly varied, spanning topics from news to finance to science, and have binary or multiple-choice (MCQ) output labels. It is difficult to draw a consistent relationship between test tasks and the finetuning datasets that lead to their largest improvement \textbf{(Best dataset)}. For example, \texttt{cluster 7} is a dataset about web cookies, yet it is the most helpful finetuning dataset for both \texttt{ag\_news} and \texttt{amazon\_polarity} which are news classification and sentiment classification tasks respectively.
Our examples of unintuitive task transfer contradict prior work that suggest domain similarity is key for successful task transfer \cite{gururangan2020don}. \citet{vu2020exploring} observed that ``Out-of-class transfer succeeds in many cases, some of which are unintuitive.'' In our experiments, unintuitive transfer appears to be the norm rather than the exception.


\subsection{Do different datasets lead to different improvements?}

\begin{figure}[h]
    \centering
    \includegraphics[width=.48\textwidth]{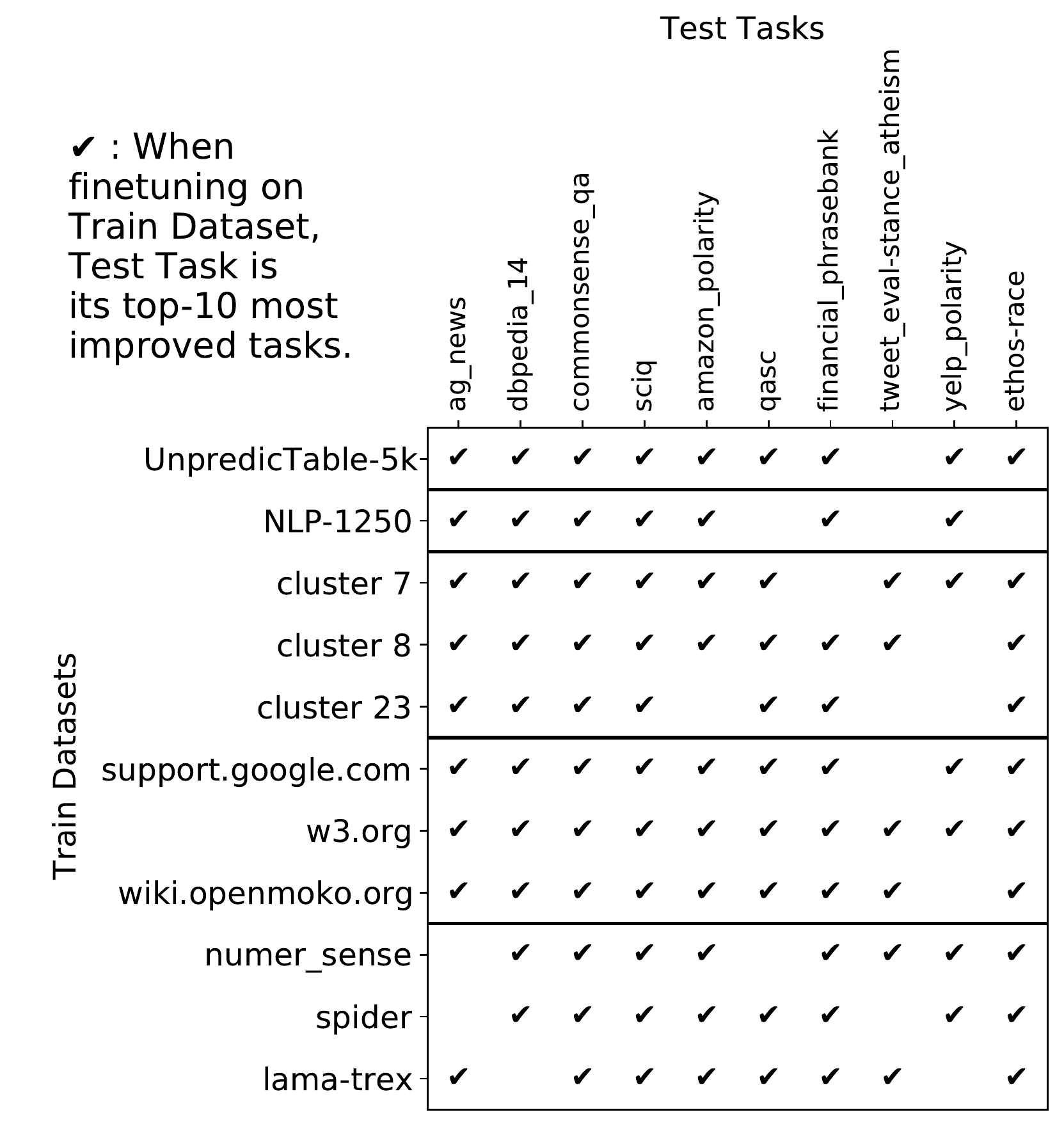}
    \caption{Finetuning on different datasets leads to broadly similar improvements. For example, finetuning on \texttt{wiki.openmoko.org} (software documentation) and \texttt{lama-trex} (factual knowledge) lead to 8 of the same test tasks being in their respective top-10 most-improved test tasks. (Out of 52 total test tasks)}
    \label{fig:top10_tasks}
\end{figure}

We wish to understand if finetuning on different datasets lead to different test task improvements. Fig. \ref{fig:top10_tasks} illustrates that the same set of 10 test tasks make up the majority of the top-10 improving test tasks for each of our best training datasets (the top-performing datasets for each category in Fig. \ref{sec:distributions}). For example, training on \texttt{wiki.openmoko.org} (software documentation) leads to strong improvements on broadly similar tasks as training on \texttt{lama-trex} (factual knowledge). This suggests that the improvements learned from these highly different training datasets are domain-agnostic. However, it remains unclear why these improvements can be learned from these particular training datasets but not others, and why these particular test tasks benefit most from the improvements.

\section{Related Work}


We focus on the FSL setting where a small number of training samples are available to learn a given task. Pretrained LMs can learn from few-shot examples in-context \cite{brown2020language, scao2021many} but have weaknesses including prompt sensitivity \cite{lu2021fantastically, perez2021true} and miscalibration \cite{zhao2021calibrate}. \citet{min2021metaicl} and \citet{chen2021meta} adapt to the FSL setting by fine-tuning LMs to predict the target given few-shot examples in the prompt. This improves FSL performance and reduces sensitivity to example ordering and example choice. We adopt MetaICL \cite{min2021metaicl} as the training method for our main experiments and support our results with additional few-shot benchmarks, CrossFit \cite{ye2021crossfit} and FLEX \cite{bragg2021flex}.

Our work also connects with other work in domain adaptation. \citet{gururangan2020don} show that fine-tuning on domains related to the downstream task leads to performance gains. Recent examples of successful domain adaptation include \citet{chen2021evaluating} for coding tasks and \citet{lewkowycz2022solving} for mathematics tasks. \citet{solaiman2021process} demonstrated this for less explicit domains, finetuning LMs on values-aligned text to generate text in accordance with intrinsic human values. In contrast, we show that LMs can be finetuned on unrelated domains yet improve on the downstream task. Other work in adaptation focus on specific task formats: \citet{khashabi2020unifiedqa, huber2021ccqa, zhong2021meta} convert broad NLP tasks into question-answering tasks and finetune to excel at question-answering; \citet{zhong2021adapting} finetunes models to perform classification tasks; \citet{gao2020making} introduce prompt templates and finetune the model to perform tasks within those templates.
More generally, LMs have been finetuned to follow instructions \cite{ouyang2022training, wei2021finetuned} which allows for more diverse tasks in various formats.
Our adaptation to FSL can be seen as adaptation to the FSL prompt format, though the tasks themselves can be diverse in domain and structure.
Multi-task literature have shown that training on a wide variety of tasks improves generalization to new task settings, which motivates our exploration of a large scale few-shot task dataset. \citet{sanh2021multitask, aribandi2021ext5, mishra2021cross, aghajanyan2021muppet, padmakumar2022exploring} demonstrate that increasing the number of tasks for multi-task training improves generalization in the zero-shot setting. \citet{xu2022zeroprompt, wang2022benchmarking} have extended this result to more than 1,000 tasks. We were inspired by these results to obtain a training dataset with 100x more tasks, but found diverse task datasets less helpful than certain narrow datasets. \citet{padmakumar2022exploring} showed that a poor choice of training task can negatively impact downstream performance, which could explain why mixing diverse tasks underperform well-chosen narrow tasks. This begs the question of how to select training datasets to improve downstream task performance. \citet{vu2020exploring} show that domain similarity can be used as a predictor for successful transfer. Our results highlight a gap in this explanation, and suggest that there may be some domain-agnostic improvements to be gained from training tasks that are unrelated to the test tasks. Other attempts to understand the effect of training datasets on FSL also struggle to uncover clean rules; this includes analyses of pretraining datasets \cite{shin2022effect}, varying datasets alongside model architectures \cite{chan2022data}, and influence functions to trace gradient updates to training datapoints \cite{akyurek2022tracing}.

Our use of structured datasets to generate training tasks is inspired by other work, though others have focused on a limited set of task types. \citet{yoran2021turning} also turn tables into tasks, using handwritten templates to extract question-answer pairs from tables. \citet{aghajanyan2021htlm} train LMs to predict masked spans in HTML webpages, then use HTML markup to prompt language models to do summarization and classification tasks. \citet{chen2022improving} transform ordinary (non-table) text into sentence completion, masked phrase prediction, and classification tasks. In contrast, our approach captures any tasks that occur in tables.


\section{Conclusion}

We produced \texttt{UnpredicTable}, a dataset of 413,299 diverse few-shot learning tasks from internet tables. Finetuning on \texttt{UnpredicTable} improves the FSL ability of LMs. However, the size of our dataset is not the key factor in its success. We find that certain narrow datasets (even ones made of trivia) are even more helpful than diverse, curated NLP datasets. Finetuning on these narrow datasets leads to strong improvements on the same test tasks as finetuning on diverse, curated NLP datasets. This suggests that finetuning on these datasets cause domain-agnostic FSL gains, though we were unable to find clear patterns to explain why this happens for some data and not others.
Our results question common wisdom that task diversity is necessary for adapting LMs to FSL. 
We hope our work spurs investigation on what data causes few-shot learning to emerge, both to develop better datasets and to better understand how training data leads to unexpected behaviors or failures.







\section{Acknowledgements}
We are grateful to Owain Evans, Mary Phuong, Seraphina Nix, and Sam Bowman for helpful conversations and feedback, as well as to Kath Lupante for task quality annotations.
We thank Open Philanthropy for funding that enabled this research.
Ethan Perez thanks the National Science Foundation and Open Philanthropy for fellowship support.

\bibliography{references}

\begin{thebibliography}{139}
\expandafter\ifx\csname natexlab\endcsname\relax\def\natexlab#1{#1}\fi

\bibitem[{Aghajanyan et~al.(2021{\natexlab{a}})Aghajanyan, Gupta, Shrivastava,
  Chen, Zettlemoyer, and Gupta}]{aghajanyan2021muppet}
Armen Aghajanyan, Anchit Gupta, Akshat Shrivastava, Xilun Chen, Luke
  Zettlemoyer, and Sonal Gupta. 2021{\natexlab{a}}.
\newblock Muppet: Massive multi-task representations with pre-finetuning.
\newblock \emph{arXiv preprint arXiv:2101.11038}.

\bibitem[{Aghajanyan et~al.(2021{\natexlab{b}})Aghajanyan, Okhonko, Lewis,
  Joshi, Xu, Ghosh, and Zettlemoyer}]{aghajanyan2021htlm}
Armen Aghajanyan, Dmytro Okhonko, Mike Lewis, Mandar Joshi, Hu~Xu, Gargi Ghosh,
  and Luke Zettlemoyer. 2021{\natexlab{b}}.
\newblock Htlm: Hyper-text pre-training and prompting of language models.
\newblock \emph{arXiv preprint arXiv:2107.06955}.

\bibitem[{Aky{\"u}rek et~al.(2022)Aky{\"u}rek, Bolukbasi, Liu, Xiong, Tenney,
  Andreas, and Guu}]{akyurek2022tracing}
Ekin Aky{\"u}rek, Tolga Bolukbasi, Frederick Liu, Binbin Xiong, Ian Tenney,
  Jacob Andreas, and Kelvin Guu. 2022.
\newblock Tracing knowledge in language models back to the training data.
\newblock \emph{arXiv preprint arXiv:2205.11482}.

\bibitem[{Almeida et~al.(2011)Almeida, Hidalgo, and Yamakami}]{sms_spam}
Tiago~A. Almeida, Jos\'{e} Mar\'{\i}a~G. Hidalgo, and Akebo Yamakami. 2011.
\newblock Contributions to the study of sms spam filtering: New collection and
  results.
\newblock In \emph{Proceedings of the 11th ACM Symposium on Document
  Engineering}.

\bibitem[{Aribandi et~al.(2021)Aribandi, Tay, Schuster, Rao, Zheng, Mehta,
  Zhuang, Tran, Bahri, Ni et~al.}]{aribandi2021ext5}
Vamsi Aribandi, Yi~Tay, Tal Schuster, Jinfeng Rao, Huaixiu~Steven Zheng,
  Sanket~Vaibhav Mehta, Honglei Zhuang, Vinh~Q Tran, Dara Bahri, Jianmo Ni,
  et~al. 2021.
\newblock Ext5: Towards extreme multi-task scaling for transfer learning.
\newblock \emph{arXiv preprint arXiv:2111.10952}.

\bibitem[{Bar-Haim et~al.(2006)Bar-Haim, Dagan, Dolan, Ferro, Giampiccolo,
  Magnini, and Szpektor}]{bar2006second}
Roy Bar-Haim, Ido Dagan, Bill Dolan, Lisa Ferro, Danilo Giampiccolo, Bernardo
  Magnini, and Idan Szpektor. 2006.
\newblock The second pascal recognising textual entailment challenge.
\newblock In \emph{Proceedings of the second PASCAL challenges workshop on
  recognising textual entailment}.

\bibitem[{Barbieri et~al.(2020)Barbieri, Camacho-Collados, Espinosa~Anke, and
  Neves}]{barbieri-etal-2020-tweeteval}
Francesco Barbieri, Jose Camacho-Collados, Luis Espinosa~Anke, and Leonardo
  Neves. 2020.
\newblock {T}weet{E}val: Unified benchmark and comparative evaluation for tweet
  classification.
\newblock In \emph{Findings of the Association for Computational Linguistics:
  EMNLP 2020}.

\bibitem[{Bentivogli et~al.(2009)Bentivogli, Clark, Dagan, and
  Giampiccolo}]{bentivogli2009fifth}
Luisa Bentivogli, Peter Clark, Ido Dagan, and Danilo Giampiccolo. 2009.
\newblock The fifth pascal recognizing textual entailment challenge.
\newblock In \emph{TAC}.

\bibitem[{Berant et~al.(2013)Berant, Chou, Frostig, and
  Liang}]{berant-etal-2013-semantic}
Jonathan Berant, Andrew Chou, Roy Frostig, and Percy Liang. 2013.
\newblock Semantic parsing on {F}reebase from question-answer pairs.
\newblock In \emph{EMNLP}.

\bibitem[{Bhagavatula et~al.(2020)Bhagavatula, Bras, Malaviya, Sakaguchi,
  Holtzman, Rashkin, Downey, tau Yih, and Choi}]{bhagavatula2020abductive}
Chandra Bhagavatula, Ronan~Le Bras, Chaitanya Malaviya, Keisuke Sakaguchi, Ari
  Holtzman, Hannah Rashkin, Doug Downey, Wen tau Yih, and Yejin Choi. 2020.
\newblock Abductive commonsense reasoning.
\newblock In \emph{ICLR}.

\bibitem[{Bisk et~al.(2020)Bisk, Zellers, Bras, Gao, and Choi}]{bisk2019piqa}
Yonatan Bisk, Rowan Zellers, Ronan~Le Bras, Jianfeng Gao, and Yejin Choi. 2020.
\newblock Piqa: Reasoning about physical commonsense in natural language.
\newblock In \emph{AAAI}.

\bibitem[{Boratko et~al.(2020)Boratko, Li, O{'}Gorman, Das, Le, and
  McCallum}]{boratko-etal-2020-protoqa}
Michael Boratko, Xiang Li, Tim O{'}Gorman, Rajarshi Das, Dan Le, and Andrew
  McCallum. 2020.
\newblock {P}roto{QA}: A question answering dataset for prototypical
  common-sense reasoning.
\newblock In \emph{EMNLP}.

\bibitem[{Bragg et~al.(2021)Bragg, Cohan, Lo, and Beltagy}]{bragg2021flex}
Jonathan Bragg, Arman Cohan, Kyle Lo, and Iz~Beltagy. 2021.
\newblock Flex: Unifying evaluation for few-shot nlp.
\newblock \emph{Advances in Neural Information Processing Systems},
  34:15787--15800.

\bibitem[{Brown et~al.(2020)Brown, Mann, Ryder, Subbiah, Kaplan, Dhariwal,
  Neelakantan, Shyam, Sastry, Askell et~al.}]{brown2020language}
Tom Brown, Benjamin Mann, Nick Ryder, Melanie Subbiah, Jared~D Kaplan, Prafulla
  Dhariwal, Arvind Neelakantan, Pranav Shyam, Girish Sastry, Amanda Askell,
  et~al. 2020.
\newblock Language models are few-shot learners.
\newblock \emph{Advances in neural information processing systems},
  33:1877--1901.

\bibitem[{Chan et~al.(2022)Chan, Santoro, Lampinen, Wang, Singh, Richemond,
  McClelland, and Hill}]{chan2022data}
Stephanie~CY Chan, Adam Santoro, Andrew~K Lampinen, Jane~X Wang, Aaditya Singh,
  Pierre~H Richemond, Jay McClelland, and Felix Hill. 2022.
\newblock Data distributional properties drive emergent few-shot learning in
  transformers.
\newblock \emph{arXiv preprint arXiv:2205.05055}.

\bibitem[{Chatterjee et~al.(2019)Chatterjee, Narahari, Joshi, and
  Agrawal}]{chatterjee-etal-2019-semeval}
Ankush Chatterjee, Kedhar~Nath Narahari, Meghana Joshi, and Puneet Agrawal.
  2019.
\newblock {S}em{E}val-2019 task 3: {E}mo{C}ontext contextual emotion detection
  in text.
\newblock In \emph{Proceedings of the 13th International Workshop on Semantic
  Evaluation}.

\bibitem[{Chen et~al.(2021{\natexlab{a}})Chen, Tworek, Jun, Yuan, Pinto,
  Kaplan, Edwards, Burda, Joseph, Brockman et~al.}]{chen2021evaluating}
Mark Chen, Jerry Tworek, Heewoo Jun, Qiming Yuan, Henrique Ponde de~Oliveira
  Pinto, Jared Kaplan, Harri Edwards, Yuri Burda, Nicholas Joseph, Greg
  Brockman, et~al. 2021{\natexlab{a}}.
\newblock Evaluating large language models trained on code.
\newblock \emph{arXiv preprint arXiv:2107.03374}.

\bibitem[{Chen et~al.(2019)Chen, D{'}Arcy, Liu, Fernandez, and
  Downey}]{chen-etal-2019-codah}
Michael Chen, Mike D{'}Arcy, Alisa Liu, Jared Fernandez, and Doug Downey. 2019.
\newblock {CODAH}: An adversarially-authored question answering dataset for
  common sense.
\newblock In \emph{Proceedings of the 3rd Workshop on Evaluating Vector Space
  Representations for {NLP}}.

\bibitem[{Chen et~al.(2022)Chen, Du, Pasunuru, Mihaylov, Iyer, Stoyanov, and
  Kozareva}]{chen2022improving}
Mingda Chen, Jingfei Du, Ramakanth Pasunuru, Todor Mihaylov, Srini Iyer,
  Veselin Stoyanov, and Zornitsa Kozareva. 2022.
\newblock Improving in-context few-shot learning via self-supervised training.
\newblock \emph{arXiv preprint arXiv:2205.01703}.

\bibitem[{Chen et~al.(2020)Chen, Wang, Chen, Zhang, Wang, Li, Zhou, and
  Wang}]{Chen2020TabFact}
Wenhu Chen, Hongmin Wang, Jianshu Chen, Yunkai Zhang, Hong Wang, Shiyang Li,
  Xiyou Zhou, and William~Yang Wang. 2020.
\newblock Tabfact: A large-scale dataset for table-based fact verification.
\newblock In \emph{ICLR}.

\bibitem[{Chen et~al.(2021{\natexlab{b}})Chen, Zhong, Zha, Karypis, and
  He}]{chen2021meta}
Yanda Chen, Ruiqi Zhong, Sheng Zha, George Karypis, and He~He.
  2021{\natexlab{b}}.
\newblock Meta-learning via language model in-context tuning.
\newblock \emph{arXiv preprint arXiv:2110.07814}.

\bibitem[{Clark et~al.(2019)Clark, Lee, Chang, Kwiatkowski, Collins, and
  Toutanova}]{clark-etal-2019-boolq}
Christopher Clark, Kenton Lee, Ming-Wei Chang, Tom Kwiatkowski, Michael
  Collins, and Kristina Toutanova. 2019.
\newblock {B}ool{Q}: Exploring the surprising difficulty of natural yes/no
  questions.
\newblock In \emph{NAACL-HLT}.

\bibitem[{Clark et~al.(2018)Clark, Cowhey, Etzioni, Khot, Sabharwal, Schoenick,
  and Tafjord}]{Clark2018ThinkYH}
Peter Clark, Isaac Cowhey, Oren Etzioni, Tushar Khot, Ashish Sabharwal, Carissa
  Schoenick, and Oyvind Tafjord. 2018.
\newblock Think you have solved question answering? try arc, the ai2 reasoning
  challenge.
\newblock \emph{arXiv preprint arXiv:1803.05457}.

\bibitem[{Dagan et~al.(2005)Dagan, Glickman, and Magnini}]{dagan2005pascal}
Ido Dagan, Oren Glickman, and Bernardo Magnini. 2005.
\newblock The pascal recognising textual entailment challenge.
\newblock In \emph{Machine Learning Challenges Workshop}.

\bibitem[{Dasigi et~al.(2019)Dasigi, Liu, Marasovi{\'c}, Smith, and
  Gardner}]{dasigi-etal-2019-quoref}
Pradeep Dasigi, Nelson~F. Liu, Ana Marasovi{\'c}, Noah~A. Smith, and Matt
  Gardner. 2019.
\newblock {Q}uoref: A reading comprehension dataset with questions requiring
  coreferential reasoning.
\newblock In \emph{EMNLP}.

\bibitem[{Davidson et~al.(2017)Davidson, Warmsley, Macy, and
  Weber}]{hateoffensive}
Thomas Davidson, Dana Warmsley, Michael Macy, and Ingmar Weber. 2017.
\newblock Automated hate speech detection and the problem of offensive
  language.
\newblock In \emph{Proceedings of the 11th International AAAI Conference on Web
  and Social Media}.

\bibitem[{de~Gibert et~al.(2018)de~Gibert, Perez, Garc{\'\i}a-Pablos, and
  Cuadros}]{gibert2018hate}
Ona de~Gibert, Naiara Perez, Aitor Garc{\'\i}a-Pablos, and Montse Cuadros.
  2018.
\newblock {Hate Speech Dataset from a White Supremacy Forum}.
\newblock In \emph{Proceedings of the 2nd Workshop on Abusive Language Online
  ({ALW}2)}.

\bibitem[{de~Marneffe et~al.(2019)de~Marneffe, Simons, and
  Tonhauser}]{Marneffe_Simons_Tonhauser_2019}
Marie-Catherine de~Marneffe, Mandy Simons, and Judith Tonhauser. 2019.
\newblock The commitmentbank: Investigating projection in naturally occurring
  discourse.
\newblock \emph{Proceedings of Sinn und Bedeutung}.

\bibitem[{Diggelmann et~al.(2020)Diggelmann, Boyd-Graber, Bulian, Ciaramita,
  and Leippold}]{Diggelmann2020CLIMATEFEVERAD}
T.~Diggelmann, Jordan~L. Boyd-Graber, Jannis Bulian, Massimiliano Ciaramita,
  and Markus Leippold. 2020.
\newblock Climate-fever: A dataset for verification of real-world climate
  claims.
\newblock \emph{ArXiv}.

\bibitem[{Dolan and Brockett(2005)}]{dolan-brockett-2005-automatically}
William~B. Dolan and Chris Brockett. 2005.
\newblock Automatically constructing a corpus of sentential paraphrases.
\newblock In \emph{Proceedings of the Third International Workshop on
  Paraphrasing ({IWP}2005)}.

\bibitem[{Dua et~al.(2019)Dua, Wang, Dasigi, Stanovsky, Singh, and
  Gardner}]{dua-etal-2019-drop}
Dheeru Dua, Yizhong Wang, Pradeep Dasigi, Gabriel Stanovsky, Sameer Singh, and
  Matt Gardner. 2019.
\newblock {DROP}: {A} reading comprehension benchmark requiring discrete
  reasoning over paragraphs.
\newblock In \emph{NAACL}.

\bibitem[{Dunn et~al.(2017)Dunn, Sagun, Higgins, G{\"u}ney, Cirik, and
  Cho}]{Dunn2017SearchQAAN}
Matthew Dunn, Levent Sagun, Mike Higgins, V.~U. G{\"u}ney, Volkan Cirik, and
  Kyunghyun Cho. 2017.
\newblock Searchqa: A new q\&a dataset augmented with context from a search
  engine.
\newblock \emph{arXiv preprint arXiv:1704.05179}.

\bibitem[{Elsahar et~al.(2018)Elsahar, Vougiouklis, Remaci, Gravier, Hare,
  Laforest, and Simperl}]{elsahar-etal-2018-rex}
Hady Elsahar, Pavlos Vougiouklis, Arslen Remaci, Christophe Gravier, Jonathon
  Hare, Frederique Laforest, and Elena Simperl. 2018.
\newblock T-{RE}x: A large scale alignment of natural language with knowledge
  base triples.
\newblock In \emph{LREC}.

\bibitem[{Faruqui and Das(2018)}]{faruqui-das-2018-identifying}
Manaal Faruqui and Dipanjan Das. 2018.
\newblock Identifying well-formed natural language questions.
\newblock In \emph{EMNLP}.

\bibitem[{Gao et~al.(2020)Gao, Fisch, and Chen}]{gao2020making}
Tianyu Gao, Adam Fisch, and Danqi Chen. 2020.
\newblock Making pre-trained language models better few-shot learners.
\newblock \emph{arXiv preprint arXiv:2012.15723}.

\bibitem[{Giampiccolo et~al.(2007)Giampiccolo, Magnini, Dagan, and
  Dolan}]{giampiccolo2007third}
Danilo Giampiccolo, Bernardo Magnini, Ido Dagan, and Bill Dolan. 2007.
\newblock The third pascal recognizing textual entailment challenge.
\newblock In \emph{Proceedings of the ACL-PASCAL workshop on textual entailment
  and paraphrasing}.

\bibitem[{Gordon et~al.(2012)Gordon, Kozareva, and
  Roemmele}]{gordon-etal-2012-semeval}
Andrew Gordon, Zornitsa Kozareva, and Melissa Roemmele. 2012.
\newblock {S}em{E}val-2012 task 7: Choice of plausible alternatives: An
  evaluation of commonsense causal reasoning.
\newblock In \emph{The First Joint Conference on Lexical and Computational
  Semantics ({S}em{E}val)}.

\bibitem[{Gurulingappa et~al.(2012)Gurulingappa, Rajput, Roberts, Fluck,
  Hofmann-Apitius, and Toldo}]{GURULINGAPPA2012885}
Harsha Gurulingappa, Abdul~Mateen Rajput, Angus Roberts, Juliane Fluck, Martin
  Hofmann-Apitius, and Luca Toldo. 2012.
\newblock Development of a benchmark corpus to support the automatic extraction
  of drug-related adverse effects from medical case reports.
\newblock \emph{Journal of Biomedical Informatics}.

\bibitem[{Gururangan et~al.(2020)Gururangan, Marasovi{\'c}, Swayamdipta, Lo,
  Beltagy, Downey, and Smith}]{gururangan2020don}
Suchin Gururangan, Ana Marasovi{\'c}, Swabha Swayamdipta, Kyle Lo, Iz~Beltagy,
  Doug Downey, and Noah~A Smith. 2020.
\newblock Don't stop pretraining: adapt language models to domains and tasks.
\newblock \emph{arXiv preprint arXiv:2004.10964}.

\bibitem[{He et~al.(2015)He, Lewis, and Zettlemoyer}]{he-etal-2015-question}
Luheng He, Mike Lewis, and Luke Zettlemoyer. 2015.
\newblock Question-answer driven semantic role labeling: Using natural language
  to annotate natural language.
\newblock In \emph{Proceedings of the 2015 Conference on Empirical Methods in
  Natural Language Processing}.

\bibitem[{Hoffart et~al.(2011)Hoffart, Yosef, Bordino, F{\"u}rstenau, Pinkal,
  Spaniol, Taneva, Thater, and Weikum}]{hoffart-etal-2011-robust}
Johannes Hoffart, Mohamed~Amir Yosef, Ilaria Bordino, Hagen F{\"u}rstenau,
  Manfred Pinkal, Marc Spaniol, Bilyana Taneva, Stefan Thater, and Gerhard
  Weikum. 2011.
\newblock Robust disambiguation of named entities in text.
\newblock In \emph{EMNLP}.

\bibitem[{Hovy et~al.(2001)Hovy, Gerber, Hermjakob, Lin, and
  Ravichandran}]{hovy-etal-2001-toward}
Eduard Hovy, Laurie Gerber, Ulf Hermjakob, Chin-Yew Lin, and Deepak
  Ravichandran. 2001.
\newblock Toward semantics-based answer pinpointing.
\newblock In \emph{Proceedings of the First International Conference on Human
  Language Technology Research}.

\bibitem[{Huang et~al.(2019)Huang, Le~Bras, Bhagavatula, and
  Choi}]{huang-etal-2019-cosmos}
Lifu Huang, Ronan Le~Bras, Chandra Bhagavatula, and Yejin Choi. 2019.
\newblock Cosmos {QA}: Machine reading comprehension with contextual
  commonsense reasoning.
\newblock In \emph{EMNLP}.

\bibitem[{Huber et~al.(2021)Huber, Aghajanyan, O{\u{g}}uz, Okhonko, Yih, Gupta,
  and Chen}]{huber2021ccqa}
Patrick Huber, Armen Aghajanyan, Barlas O{\u{g}}uz, Dmytro Okhonko, Wen-tau
  Yih, Sonal Gupta, and Xilun Chen. 2021.
\newblock Ccqa: A new web-scale question answering dataset for model
  pre-training.
\newblock \emph{arXiv preprint arXiv:2110.07731}.

\bibitem[{Jiang et~al.(2019)Jiang, Wu, and Jiang}]{jiang-etal-2019-freebaseqa}
Kelvin Jiang, Dekun Wu, and Hui Jiang. 2019.
\newblock {F}reebase{QA}: A new factoid {QA} data set matching trivia-style
  question-answer pairs with {F}reebase.
\newblock In \emph{NAACL-HLT}.

\bibitem[{Khashabi et~al.(2018)Khashabi, Chaturvedi, Roth, Upadhyay, and
  Roth}]{khashabi-etal-2018-looking}
Daniel Khashabi, Snigdha Chaturvedi, Michael Roth, Shyam Upadhyay, and Dan
  Roth. 2018.
\newblock Looking beyond the surface: A challenge set for reading comprehension
  over multiple sentences.
\newblock In \emph{NAACL-HLT}.

\bibitem[{Khashabi et~al.(2020)Khashabi, Min, Khot, Sabharwal, Tafjord, Clark,
  and Hajishirzi}]{khashabi2020unifiedqa}
Daniel Khashabi, Sewon Min, Tushar Khot, Ashish Sabharwal, Oyvind Tafjord,
  Peter Clark, and Hannaneh Hajishirzi. 2020.
\newblock Unifiedqa: Crossing format boundaries with a single qa system.
\newblock \emph{arXiv preprint arXiv:2005.00700}.

\bibitem[{Khot et~al.(2019)Khot, Clark, Guerquin, Jansen, and
  Sabharwal}]{khot2019qasc}
Tushar Khot, Peter Clark, Michal Guerquin, Peter Jansen, and Ashish Sabharwal.
  2019.
\newblock {QASC}: A dataset for question answering via sentence composition.
\newblock In \emph{AAAI}.

\bibitem[{Khot et~al.(2020)Khot, Clark, Guerquin, Jansen, and
  Sabharwal}]{Khot_Clark_Guerquin_Jansen_Sabharwal_2020}
Tushar Khot, Peter Clark, Michal Guerquin, Peter Jansen, and Ashish Sabharwal.
  2020.
\newblock Qasc: A dataset for question answering via sentence composition.
\newblock In \emph{AAAI}.

\bibitem[{Khot et~al.(2018)Khot, Sabharwal, and Clark}]{scitail}
Tushar Khot, Ashish Sabharwal, and Peter Clark. 2018.
\newblock Scitail: A textual entailment dataset from science question
  answering.
\newblock In \emph{AAAI}.

\bibitem[{Kocisk{\'y} et~al.(2018)Kocisk{\'y}, Schwarz, Blunsom, Dyer, Hermann,
  Melis, and Grefenstette}]{kocisky-etal-2018-narrativeqa}
Tom{\'a}s Kocisk{\'y}, Jonathan Schwarz, Phil Blunsom, Chris Dyer, Karl~Moritz
  Hermann, G{\'a}bor Melis, and Edward Grefenstette. 2018.
\newblock The narrativeqa reading comprehension challenge.
\newblock \emph{TACL}.

\bibitem[{Kotonya and Toni(2020)}]{kotonya-toni-2020-explainable-automated}
Neema Kotonya and Francesca Toni. 2020.
\newblock Explainable automated fact-checking for public health claims.
\newblock In \emph{EMNLP}.

\bibitem[{Kwiatkowski et~al.(2019)Kwiatkowski, Palomaki, Redfield, Collins,
  Parikh, Alberti, Epstein, Polosukhin, Devlin, Lee, Toutanova, Jones, Kelcey,
  Chang, Dai, Uszkoreit, Le, and Petrov}]{kwiatkowski-etal-2019-natural}
Tom Kwiatkowski, Jennimaria Palomaki, Olivia Redfield, Michael Collins,
  Ankur~P. Parikh, Chris Alberti, Danielle Epstein, Illia Polosukhin, Jacob
  Devlin, Kenton Lee, Kristina Toutanova, Llion Jones, Matthew Kelcey, Ming-Wei
  Chang, Andrew~M. Dai, Jakob Uszkoreit, Quoc Le, and Slav Petrov. 2019.
\newblock Natural questions: A benchmark for question answering research.
\newblock \emph{TACL}.

\bibitem[{Lai et~al.(2017)Lai, Xie, Liu, Yang, and Hovy}]{lai-etal-2017-race}
Guokun Lai, Qizhe Xie, Hanxiao Liu, Yiming Yang, and Eduard~H. Hovy. 2017.
\newblock {RACE}: {L}arge-scale reading comprehension dataset from
  examinations.
\newblock In \emph{EMNLP}.

\bibitem[{Lehmann et~al.(2015)Lehmann, Isele, Jakob, Jentzsch, Kontokostas,
  Mendes, Hellmann, Morsey, van Kleef, Auer, and Bizer}]{Lehmann2015DBpediaA}
Jens Lehmann, Robert Isele, Max Jakob, Anja Jentzsch, D.~Kontokostas, Pablo~N.
  Mendes, Sebastian Hellmann, M.~Morsey, Patrick van Kleef, S.~Auer, and
  C.~Bizer. 2015.
\newblock Dbpedia - a large-scale, multilingual knowledge base extracted from
  wikipedia.
\newblock \emph{Semantic Web}.

\bibitem[{Levesque et~al.(2012)Levesque, Davis, and
  Morgenstern}]{levesque2012winograd}
Hector~J. Levesque, Ernest Davis, and Leora Morgenstern. 2012.
\newblock The winograd schema challenge.
\newblock In \emph{Proceedings of the Thirteenth International Conference on
  Principles of Knowledge Representation and Reasoning}.

\bibitem[{Levy et~al.(2017)Levy, Seo, Choi, and
  Zettlemoyer}]{levy-etal-2017-zero}
Omer Levy, Minjoon Seo, Eunsol Choi, and Luke Zettlemoyer. 2017.
\newblock Zero-shot relation extraction via reading comprehension.
\newblock In \emph{{C}o{NLL}}.

\bibitem[{Lewis et~al.(2019)Lewis, Liu, Goyal, Ghazvininejad, Mohamed, Levy,
  Stoyanov, and Zettlemoyer}]{lewis2019bart}
Mike Lewis, Yinhan Liu, Naman Goyal, Marjan Ghazvininejad, Abdelrahman Mohamed,
  Omer Levy, Ves Stoyanov, and Luke Zettlemoyer. 2019.
\newblock Bart: Denoising sequence-to-sequence pre-training for natural
  language generation, translation, and comprehension.
\newblock \emph{arXiv preprint arXiv:1910.13461}.

\bibitem[{Lewkowycz et~al.(2022)Lewkowycz, Andreassen, Dohan, Dyer,
  Michalewski, Ramasesh, Slone, Anil, Schlag, Gutman-Solo
  et~al.}]{lewkowycz2022solving}
Aitor Lewkowycz, Anders Andreassen, David Dohan, Ethan Dyer, Henryk
  Michalewski, Vinay Ramasesh, Ambrose Slone, Cem Anil, Imanol Schlag, Theo
  Gutman-Solo, et~al. 2022.
\newblock Solving quantitative reasoning problems with language models.
\newblock \emph{arXiv preprint arXiv:2206.14858}.

\bibitem[{Li and Roth(2002)}]{li-roth-2002-learning}
Xin Li and Dan Roth. 2002.
\newblock Learning question classifiers.
\newblock In \emph{COLING}.

\bibitem[{Lin et~al.(2020)Lin, Lee, Khanna, and Ren}]{lin-etal-2020-birds}
Bill~Yuchen Lin, Seyeon Lee, Rahul Khanna, and Xiang Ren. 2020.
\newblock {B}irds have four legs?! {N}umer{S}ense: {P}robing {N}umerical
  {C}ommonsense {K}nowledge of {P}re-{T}rained {L}anguage {M}odels.
\newblock In \emph{EMNLP}.

\bibitem[{Lin et~al.(2019)Lin, Tafjord, Clark, and
  Gardner}]{lin-etal-2019-reasoning}
Kevin Lin, Oyvind Tafjord, Peter Clark, and Matt Gardner. 2019.
\newblock Reasoning over paragraph effects in situations.
\newblock In \emph{Proceedings of the 2nd Workshop on Machine Reading for
  Question Answering}.

\bibitem[{Louis et~al.(2020)Louis, Roth, and Radlinski}]{louis-etal-2020-id}
Annie Louis, Dan Roth, and Filip Radlinski. 2020.
\newblock {``}{I}{'}d rather just go to bed{''}: Understanding indirect
  answers.
\newblock In \emph{EMNLP}.

\bibitem[{Lu et~al.(2021)Lu, Bartolo, Moore, Riedel, and
  Stenetorp}]{lu2021fantastically}
Yao Lu, Max Bartolo, Alastair Moore, Sebastian Riedel, and Pontus Stenetorp.
  2021.
\newblock Fantastically ordered prompts and where to find them: Overcoming
  few-shot prompt order sensitivity.
\newblock \emph{arXiv preprint arXiv:2104.08786}.

\bibitem[{Maas et~al.(2011)Maas, Daly, Pham, Huang, Ng, and
  Potts}]{maas-etal-2011-learning}
Andrew~L. Maas, Raymond~E. Daly, Peter~T. Pham, Dan Huang, Andrew~Y. Ng, and
  Christopher Potts. 2011.
\newblock Learning word vectors for sentiment analysis.
\newblock In \emph{Proceedings of the 49th Annual Meeting of the Association
  for Computational Linguistics: Human Language Technologies}.

\bibitem[{Malo et~al.(2014)Malo, Sinha, Korhonen, Wallenius, and
  Takala}]{financial-phrasebank}
Pekka Malo, Ankur Sinha, Pekka Korhonen, Jyrki Wallenius, and Pyry Takala.
  2014.
\newblock Good debt or bad debt: Detecting semantic orientations in economic
  texts.
\newblock \emph{J. Assoc. Inf. Sci. Technol.}

\bibitem[{Marelli et~al.(2014)Marelli, Menini, Baroni, Bentivogli, Bernardi,
  and Zamparelli}]{marelli-etal-2014-sick}
Marco Marelli, Stefano Menini, Marco Baroni, Luisa Bentivogli, Raffaella
  Bernardi, and Roberto Zamparelli. 2014.
\newblock A {SICK} cure for the evaluation of compositional distributional
  semantic models.
\newblock In \emph{LREC}.

\bibitem[{Mathew et~al.(2020)Mathew, Saha, Yimam, Biemann, Goyal, and
  Mukherjee}]{mathew2020hatexplain}
Binny Mathew, Punyajoy Saha, Seid~Muhie Yimam, Chris Biemann, Pawan Goyal, and
  Animesh Mukherjee. 2020.
\newblock Hatexplain: A benchmark dataset for explainable hate speech
  detection.
\newblock \emph{arXiv preprint arXiv:2012.10289}.

\bibitem[{McAuley and Leskovec(2013)}]{McAuley2013HiddenFA}
Julian McAuley and J.~Leskovec. 2013.
\newblock Hidden factors and hidden topics: understanding rating dimensions
  with review text.
\newblock \emph{Proceedings of the 7th ACM conference on Recommender systems}.

\bibitem[{McCreery et~al.(2020)McCreery, Katariya, Kannan, Chablani, and
  Amatriain}]{medical-qqp}
Clara~H. McCreery, Namit Katariya, Anitha Kannan, Manish Chablani, and Xavier
  Amatriain. 2020.
\newblock Effective transfer learning for identifying similar questions:
  Matching user questions to covid-19 faqs.
\newblock In \emph{Proceedings of the 26th ACM SIGKDD International Conference
  on Knowledge Discovery \& Data Mining}.

\bibitem[{McInnes et~al.(2017)McInnes, Healy, and Astels}]{mcinnes2017hdbscan}
Leland McInnes, John Healy, and Steve Astels. 2017.
\newblock hdbscan: Hierarchical density based clustering.
\newblock \emph{J. Open Source Softw.}, 2(11):205.

\bibitem[{McInnes et~al.(2018)McInnes, Healy, Saul, and
  Grossberger}]{mcinnes2018umap-software}
Leland McInnes, John Healy, Nathaniel Saul, and Lukas Grossberger. 2018.
\newblock Umap: Uniform manifold approximation and projection.
\newblock \emph{The Journal of Open Source Software}, 3(29):861.

\bibitem[{Mihaylov et~al.(2018)Mihaylov, Clark, Khot, and
  Sabharwal}]{mihaylov-etal-2018-suit}
Todor Mihaylov, Peter Clark, Tushar Khot, and Ashish Sabharwal. 2018.
\newblock Can a suit of armor conduct electricity? a new dataset for open book
  question answering.
\newblock In \emph{EMNLP}.

\bibitem[{Min et~al.(2021)Min, Lewis, Zettlemoyer, and
  Hajishirzi}]{min2021metaicl}
Sewon Min, Mike Lewis, Luke Zettlemoyer, and Hannaneh Hajishirzi. 2021.
\newblock Metaicl: Learning to learn in context.
\newblock \emph{arXiv preprint arXiv:2110.15943}.

\bibitem[{Mishra et~al.(2021)Mishra, Khashabi, Baral, and
  Hajishirzi}]{mishra2021cross}
Swaroop Mishra, Daniel Khashabi, Chitta Baral, and Hannaneh Hajishirzi. 2021.
\newblock Cross-task generalization via natural language crowdsourcing
  instructions.
\newblock \emph{arXiv preprint arXiv:2104.08773}.

\bibitem[{Mollas et~al.(2020)Mollas, Chrysopoulou, Karlos, and
  Tsoumakas}]{Mollas2020ETHOSAO}
Ioannis Mollas, Zoe Chrysopoulou, Stamatis Karlos, and Grigorios Tsoumakas.
  2020.
\newblock Ethos: an online hate speech detection dataset.
\newblock \emph{arXiv preprint arXiv:2006.08328}.

\bibitem[{Nangia et~al.(2020)Nangia, Vania, Bhalerao, and
  Bowman}]{nangia-etal-2020-crows}
Nikita Nangia, Clara Vania, Rasika Bhalerao, and Samuel~R. Bowman. 2020.
\newblock {C}row{S}-pairs: A challenge dataset for measuring social biases in
  masked language models.
\newblock In \emph{EMNLP}.

\bibitem[{Napoles et~al.(2012)Napoles, Gormley, and
  Van~Durme}]{napoles-etal-2012-annotated}
Courtney Napoles, Matthew Gormley, and Benjamin Van~Durme. 2012.
\newblock Annotated {G}igaword.
\newblock In \emph{Proceedings of the Joint Workshop on Automatic Knowledge
  Base Construction and Web-scale Knowledge Extraction ({AKBC}-{WEKEX})}.

\bibitem[{Narayan et~al.(2018)Narayan, Cohen, and
  Lapata}]{narayan-etal-2018-dont}
Shashi Narayan, Shay~B. Cohen, and Mirella Lapata. 2018.
\newblock Don{'}t give me the details, just the summary! topic-aware
  convolutional neural networks for extreme summarization.
\newblock In \emph{EMNLP}.

\bibitem[{Nie et~al.(2020)Nie, Williams, Dinan, Bansal, Weston, and
  Kiela}]{nie-etal-2020-adversarial}
Yixin Nie, Adina Williams, Emily Dinan, Mohit Bansal, Jason Weston, and Douwe
  Kiela. 2020.
\newblock Adversarial {NLI}: A new benchmark for natural language
  understanding.
\newblock In \emph{ACL}.

\bibitem[{Ouyang et~al.(2022)Ouyang, Wu, Jiang, Almeida, Wainwright, Mishkin,
  Zhang, Agarwal, Slama, Ray et~al.}]{ouyang2022training}
Long Ouyang, Jeff Wu, Xu~Jiang, Diogo Almeida, Carroll~L Wainwright, Pamela
  Mishkin, Chong Zhang, Sandhini Agarwal, Katarina Slama, Alex Ray, et~al.
  2022.
\newblock Training language models to follow instructions with human feedback.
\newblock \emph{arXiv preprint arXiv:2203.02155}.

\bibitem[{Padmakumar et~al.(2022)Padmakumar, Lausen, Ballesteros, Zha, He, and
  Karypis}]{padmakumar2022exploring}
Vishakh Padmakumar, Leonard Lausen, Miguel Ballesteros, Sheng Zha, He~He, and
  George Karypis. 2022.
\newblock Exploring the role of task transferability in large-scale multi-task
  learning.
\newblock \emph{arXiv preprint arXiv:2204.11117}.

\bibitem[{Pang and Lee(2005)}]{pang-lee-2005-seeing}
Bo~Pang and Lillian Lee. 2005.
\newblock Seeing stars: Exploiting class relationships for sentiment
  categorization with respect to rating scales.
\newblock In \emph{ACL}.

\bibitem[{Pappas et~al.(2020)Pappas, Stavropoulos, Androutsopoulos, and
  McDonald}]{pappas-etal-2020-biomrc}
Dimitris Pappas, Petros Stavropoulos, Ion Androutsopoulos, and Ryan McDonald.
  2020.
\newblock {B}io{MRC}: A dataset for biomedical machine reading comprehension.
\newblock In \emph{Proceedings of the 19th SIGBioMed Workshop on Biomedical
  Language Processing}.

\bibitem[{Perez et~al.(2021)Perez, Kiela, and Cho}]{perez2021true}
Ethan Perez, Douwe Kiela, and Kyunghyun Cho. 2021.
\newblock True few-shot learning with language models.
\newblock \emph{Advances in Neural Information Processing Systems},
  34:11054--11070.

\bibitem[{Petroni et~al.(2020)Petroni, Lewis, Piktus, Rockt{\"a}schel, Wu,
  Miller, and Riedel}]{petroni2020how}
Fabio Petroni, Patrick Lewis, Aleksandra Piktus, Tim Rockt{\"a}schel, Yuxiang
  Wu, Alexander~H. Miller, and Sebastian Riedel. 2020.
\newblock How context affects language models' factual predictions.
\newblock In \emph{Automated Knowledge Base Construction}.

\bibitem[{Petroni et~al.(2019)Petroni, Rockt{\"a}schel, Riedel, Lewis, Bakhtin,
  Wu, and Miller}]{petroni-etal-2019-language}
Fabio Petroni, Tim Rockt{\"a}schel, Sebastian Riedel, Patrick Lewis, Anton
  Bakhtin, Yuxiang Wu, and Alexander Miller. 2019.
\newblock Language models as knowledge bases?
\newblock In \emph{EMNLP}.

\bibitem[{Pilehvar and
  Camacho-Collados(2019)}]{pilehvar-camacho-collados-2019-wic}
Mohammad~Taher Pilehvar and Jose Camacho-Collados. 2019.
\newblock {W}i{C}: the word-in-context dataset for evaluating context-sensitive
  meaning representations.
\newblock In \emph{NAACL-HLT}.

\bibitem[{Radford et~al.(2019)Radford, Wu, Child, Luan, Amodei, Sutskever
  et~al.}]{radford2019language}
Alec Radford, Jeffrey Wu, Rewon Child, David Luan, Dario Amodei, Ilya
  Sutskever, et~al. 2019.
\newblock Language models are unsupervised multitask learners.
\newblock \emph{OpenAI blog}, 1(8):9.

\bibitem[{Rajpurkar et~al.(2018)Rajpurkar, Jia, and
  Liang}]{rajpurkar-etal-2018-know}
Pranav Rajpurkar, Robin Jia, and Percy Liang. 2018.
\newblock Know what you don't know: Unanswerable questions for squad.
\newblock In \emph{ACL}.

\bibitem[{Rajpurkar et~al.(2016)Rajpurkar, Zhang, Lopyrev, and
  Liang}]{rajpurkar-etal-2016-squad}
Pranav Rajpurkar, Jian Zhang, Konstantin Lopyrev, and Percy Liang. 2016.
\newblock {SQuAD}: 100,000+ questions for machine comprehension of text.
\newblock In \emph{{EMNLP}}.

\bibitem[{Richardson et~al.(2013)Richardson, Burges, and
  Renshaw}]{richardson-etal-2013-mctest}
Matthew Richardson, Christopher J.~C. Burges, and Erin Renshaw. 2013.
\newblock Mctest: A challenge dataset for the open-domain machine comprehension
  of text.
\newblock In \emph{EMNLP}.

\bibitem[{Rogers et~al.(2020)Rogers, Kovaleva, Downey, and
  Rumshisky}]{Rogers_Kovaleva_Downey_Rumshisky_2020}
Anna Rogers, Olga Kovaleva, Matthew Downey, and Anna Rumshisky. 2020.
\newblock Getting closer to ai complete question answering: A set of
  prerequisite real tasks.
\newblock In \emph{AAAI}.

\bibitem[{Sakaguchi et~al.(2020{\natexlab{a}})Sakaguchi, Bras, Bhagavatula, and
  Choi}]{sakaguchi2019winogrande}
Keisuke Sakaguchi, Ronan~Le Bras, Chandra Bhagavatula, and Yejin Choi.
  2020{\natexlab{a}}.
\newblock {WINOGRANDE:} an adversarial winograd schema challenge at scale.
\newblock In \emph{AAAI}.

\bibitem[{Sakaguchi et~al.(2020{\natexlab{b}})Sakaguchi, Le~Bras, Bhagavatula,
  and Choi}]{Sakaguchi_Le_Bras_Bhagavatula_Choi_2020}
Keisuke Sakaguchi, Ronan Le~Bras, Chandra Bhagavatula, and Yejin Choi.
  2020{\natexlab{b}}.
\newblock Winogrande: An adversarial winograd schema challenge at scale.
\newblock In \emph{AAAI}.

\bibitem[{Sanh et~al.(2021)Sanh, Webson, Raffel, Bach, Sutawika, Alyafeai,
  Chaffin, Stiegler, Scao, Raja et~al.}]{sanh2021multitask}
Victor Sanh, Albert Webson, Colin Raffel, Stephen~H Bach, Lintang Sutawika,
  Zaid Alyafeai, Antoine Chaffin, Arnaud Stiegler, Teven~Le Scao, Arun Raja,
  et~al. 2021.
\newblock Multitask prompted training enables zero-shot task generalization.
\newblock \emph{arXiv preprint arXiv:2110.08207}.

\bibitem[{Sap et~al.(2019{\natexlab{a}})Sap, Rashkin, Chen, Le~Bras, and
  Choi}]{sap-etal-2019-social}
Maarten Sap, Hannah Rashkin, Derek Chen, Ronan Le~Bras, and Yejin Choi.
  2019{\natexlab{a}}.
\newblock Social {IQ}a: Commonsense reasoning about social interactions.
\newblock In \emph{EMNLP}.

\bibitem[{Sap et~al.(2019{\natexlab{b}})Sap, Rashkin, Chen, Le~Bras, and
  Choi}]{sap2019socialiqa}
Maarten Sap, Hannah Rashkin, Derek Chen, Ronan Le~Bras, and Yejin Choi.
  2019{\natexlab{b}}.
\newblock Social iqa: Commonsense reasoning about social interactions.
\newblock In \emph{EMNLP-IJCNLP}.

\bibitem[{Saravia et~al.(2018)Saravia, Liu, Huang, Wu, and
  Chen}]{saravia-etal-2018-carer}
Elvis Saravia, Hsien-Chi~Toby Liu, Yen-Hao Huang, Junlin Wu, and Yi-Shin Chen.
  2018.
\newblock {CARER}: Contextualized affect representations for emotion
  recognition.
\newblock In \emph{EMNLP}.

\bibitem[{Scao and Rush(2021)}]{scao2021many}
Teven~Le Scao and Alexander~M Rush. 2021.
\newblock How many data points is a prompt worth?
\newblock \emph{arXiv preprint arXiv:2103.08493}.

\bibitem[{Sheng and Uthus(2020)}]{sheng-uthus-2020-investigating}
Emily Sheng and David Uthus. 2020.
\newblock Investigating societal biases in a poetry composition system.
\newblock In \emph{Proceedings of the Second Workshop on Gender Bias in Natural
  Language Processing}.

\bibitem[{Shin et~al.(2022)Shin, Lee, Ahn, Kim, Kim, Kim, Cho, Lee, Park, Ha
  et~al.}]{shin2022effect}
Seongjin Shin, Sang-Woo Lee, Hwijeen Ahn, Sungdong Kim, HyoungSeok Kim, Boseop
  Kim, Kyunghyun Cho, Gichang Lee, Woomyoung Park, Jung-Woo Ha, et~al. 2022.
\newblock On the effect of pretraining corpora on in-context learning by a
  large-scale language model.
\newblock \emph{arXiv preprint arXiv:2204.13509}.

\bibitem[{Sileo et~al.(2019)Sileo, Van De~Cruys, Pradel, and
  Muller}]{sileo-etal-2019-mining}
Damien Sileo, Tim Van De~Cruys, Camille Pradel, and Philippe Muller. 2019.
\newblock Mining discourse markers for unsupervised sentence representation
  learning.
\newblock In \emph{NAACL-HLT}.

\bibitem[{Socher et~al.(2013)Socher, Perelygin, Wu, Chuang, Manning, Ng, and
  Potts}]{socher-etal-2013-recursive}
Richard Socher, Alex Perelygin, Jean Wu, Jason Chuang, Christopher~D. Manning,
  Andrew Ng, and Christopher Potts. 2013.
\newblock Recursive deep models for semantic compositionality over a sentiment
  treebank.
\newblock In \emph{EMNLP}.

\bibitem[{Solaiman and Dennison(2021)}]{solaiman2021process}
Irene Solaiman and Christy Dennison. 2021.
\newblock Process for adapting language models to society (palms) with
  values-targeted datasets.
\newblock \emph{Advances in Neural Information Processing Systems},
  34:5861--5873.

\bibitem[{Sun et~al.(2019)Sun, Yu, Chen, Yu, Choi, and
  Cardie}]{sun-etal-2019-dream}
Kai Sun, Dian Yu, Jianshu Chen, Dong Yu, Yejin Choi, and Claire Cardie. 2019.
\newblock {DREAM}: A challenge data set and models for dialogue-based reading
  comprehension.
\newblock \emph{TACL}.

\bibitem[{Tafjord et~al.(2019{\natexlab{a}})Tafjord, Clark, Gardner, Yih, and
  Sabharwal}]{Tafjord_Clark_Gardner_Yih_Sabharwal_2019}
Oyvind Tafjord, Peter Clark, Matt Gardner, Wen-tau Yih, and Ashish Sabharwal.
  2019{\natexlab{a}}.
\newblock Quarel: A dataset and models for answering questions about
  qualitative relationships.
\newblock In \emph{AAAI}.

\bibitem[{Tafjord et~al.(2019{\natexlab{b}})Tafjord, Gardner, Lin, and
  Clark}]{tafjord-etal-2019-quartz}
Oyvind Tafjord, Matt Gardner, Kevin Lin, and Peter Clark. 2019{\natexlab{b}}.
\newblock {Q}ua{RT}z: An open-domain dataset of qualitative relationship
  questions.
\newblock In \emph{EMNLP}.

\bibitem[{Talmor et~al.(2019)Talmor, Herzig, Lourie, and
  Berant}]{talmor-etal-2019-commonsenseqa}
Alon Talmor, Jonathan Herzig, Nicholas Lourie, and Jonathan Berant. 2019.
\newblock Commonsenseqa: A question answering challenge targeting commonsense
  knowledge.
\newblock In \emph{NAACL-HLT}.

\bibitem[{Tandon et~al.(2019)Tandon, Dalvi, Sakaguchi, Clark, and
  Bosselut}]{tandon-etal-2019-wiqa}
Niket Tandon, Bhavana Dalvi, Keisuke Sakaguchi, Peter Clark, and Antoine
  Bosselut. 2019.
\newblock {WIQA}: A dataset for {``}what if...{''} reasoning over procedural
  text.
\newblock In \emph{EMNLP}.

\bibitem[{Thorne et~al.(2018)Thorne, Vlachos, Christodoulopoulos, and
  Mittal}]{thorne-etal-2018-fever}
James Thorne, Andreas Vlachos, Christos Christodoulopoulos, and Arpit Mittal.
  2018.
\newblock {FEVER}: a large-scale dataset for fact extraction and
  {VER}ification.
\newblock In \emph{NAACL-HLT}.

\bibitem[{Trischler et~al.(2017)Trischler, Wang, Yuan, Harris, Sordoni,
  Bachman, and Suleman}]{trischler-etal-2017-newsqa}
Adam Trischler, Tong Wang, Xingdi Yuan, Justin Harris, Alessandro Sordoni,
  Philip Bachman, and Kaheer Suleman. 2017.
\newblock Newsqa: A machine comprehension dataset.
\newblock In \emph{Rep4NLP@ACL}.

\bibitem[{Vajjala and
  Lu{\v{c}}i{\'c}(2018)}]{vajjala-lucic-2018-onestopenglish}
Sowmya Vajjala and Ivana Lu{\v{c}}i{\'c}. 2018.
\newblock {O}ne{S}top{E}nglish corpus: A new corpus for automatic readability
  assessment and text simplification.
\newblock In \emph{Proceedings of the Thirteenth Workshop on Innovative Use of
  {NLP} for Building Educational Applications}.

\bibitem[{Vaswani et~al.(2017)Vaswani, Shazeer, Parmar, Uszkoreit, Jones,
  Gomez, Kaiser, and Polosukhin}]{vaswani2017attention}
Ashish Vaswani, Noam Shazeer, Niki Parmar, Jakob Uszkoreit, Llion Jones,
  Aidan~N Gomez, \L~ukasz Kaiser, and Illia Polosukhin. 2017.
\newblock \href
  {https://proceedings.neurips.cc/paper/2017/file/3f5ee243547dee91fbd053c1c4a845aa-Paper.pdf}
  {Attention is all you need}.
\newblock In \emph{Advances in Neural Information Processing Systems},
  volume~30. Curran Associates, Inc.

\bibitem[{Vu et~al.(2020)Vu, Wang, Munkhdalai, Sordoni, Trischler,
  Mattarella-Micke, Maji, and Iyyer}]{vu2020exploring}
Tu~Vu, Tong Wang, Tsendsuren Munkhdalai, Alessandro Sordoni, Adam Trischler,
  Andrew Mattarella-Micke, Subhransu Maji, and Mohit Iyyer. 2020.
\newblock Exploring and predicting transferability across nlp tasks.
\newblock \emph{arXiv preprint arXiv:2005.00770}.

\bibitem[{Wang(2017)}]{wang-2017-liar}
William~Yang Wang. 2017.
\newblock {``}liar, liar pants on fire{''}: A new benchmark dataset for fake
  news detection.
\newblock In \emph{ACL}.

\bibitem[{Wang et~al.(2022)Wang, Mishra, Alipoormolabashi, Kordi, Mirzaei,
  Arunkumar, Ashok, Dhanasekaran, Naik, Stap et~al.}]{wang2022benchmarking}
Yizhong Wang, Swaroop Mishra, Pegah Alipoormolabashi, Yeganeh Kordi, Amirreza
  Mirzaei, Anjana Arunkumar, Arjun Ashok, Arut~Selvan Dhanasekaran, Atharva
  Naik, David Stap, et~al. 2022.
\newblock Benchmarking generalization via in-context instructions on 1,600+
  language tasks.
\newblock \emph{arXiv preprint arXiv:2204.07705}.

\bibitem[{Warstadt et~al.(2020)Warstadt, Parrish, Liu, Mohananey, Peng, Wang,
  and Bowman}]{warstadt2019blimp}
Alex Warstadt, Alicia Parrish, Haokun Liu, Anhad Mohananey, Wei Peng, Sheng-Fu
  Wang, and Samuel~R. Bowman. 2020.
\newblock Blimp: The benchmark of linguistic minimal pairs for english.
\newblock \emph{TACL}.

\bibitem[{Warstadt et~al.(2019)Warstadt, Singh, and
  Bowman}]{warstadt-etal-2019-neural}
Alex Warstadt, Amanpreet Singh, and Samuel~R. Bowman. 2019.
\newblock Neural network acceptability judgments.
\newblock \emph{TACL}.

\bibitem[{Wei et~al.(2021)Wei, Bosma, Zhao, Guu, Yu, Lester, Du, Dai, and
  Le}]{wei2021finetuned}
Jason Wei, Maarten Bosma, Vincent~Y Zhao, Kelvin Guu, Adams~Wei Yu, Brian
  Lester, Nan Du, Andrew~M Dai, and Quoc~V Le. 2021.
\newblock Finetuned language models are zero-shot learners.
\newblock \emph{arXiv preprint arXiv:2109.01652}.

\bibitem[{Welbl et~al.(2017)Welbl, Liu, and
  Gardner}]{welbl-etal-2017-crowdsourcing}
Johannes Welbl, Nelson~F. Liu, and Matt Gardner. 2017.
\newblock Crowdsourcing multiple choice science questions.
\newblock In \emph{Proceedings of the 3rd Workshop on Noisy User-generated
  Text}.

\bibitem[{Williams et~al.(2018)Williams, Nangia, and
  Bowman}]{williams-etal-2018-broad}
Adina Williams, Nikita Nangia, and Samuel Bowman. 2018.
\newblock A broad-coverage challenge corpus for sentence understanding through
  inference.
\newblock In \emph{NAACL-HLT}.

\bibitem[{Xiong et~al.(2019)Xiong, Wu, Wang, Kulkarni, Yu, Chang, Guo, and
  Wang}]{xiong-etal-2019-tweetqa}
Wenhan Xiong, Jiawei Wu, Hong Wang, Vivek Kulkarni, Mo~Yu, Shiyu Chang,
  Xiaoxiao Guo, and William~Yang Wang. 2019.
\newblock {TWEETQA}: A social media focused question answering dataset.
\newblock In \emph{ACL}.

\bibitem[{Xu et~al.(2022)Xu, Chen, Du, Shao, Wang, Li, and
  Yang}]{xu2022zeroprompt}
Hanwei Xu, Yujun Chen, Yulun Du, Nan Shao, Yanggang Wang, Haiyu Li, and Zhilin
  Yang. 2022.
\newblock Zeroprompt: Scaling prompt-based pretraining to 1,000 tasks improves
  zero-shot generalization.
\newblock \emph{arXiv preprint arXiv:2201.06910}.

\bibitem[{Yang et~al.(2015)Yang, Yih, and Meek}]{yang-etal-2015-wikiqa}
Yi~Yang, Wen-tau Yih, and Christopher Meek. 2015.
\newblock {W}iki{QA}: A challenge dataset for open-domain question answering.
\newblock In \emph{EMNLP}.

\bibitem[{Yang et~al.(2018)Yang, Qi, Zhang, Bengio, Cohen, Salakhutdinov, and
  Manning}]{yang-etal-2018-hotpotqa}
Zhilin Yang, Peng Qi, Saizheng Zhang, Yoshua Bengio, William Cohen, Ruslan
  Salakhutdinov, and Christopher~D. Manning. 2018.
\newblock {H}otpot{QA}: A dataset for diverse, explainable multi-hop question
  answering.
\newblock In \emph{EMNLP}.

\bibitem[{Ye et~al.(2021)Ye, Lin, and Ren}]{ye2021crossfit}
Qinyuan Ye, Bill~Yuchen Lin, and Xiang Ren. 2021.
\newblock Crossfit: A few-shot learning challenge for cross-task generalization
  in nlp.
\newblock \emph{arXiv preprint arXiv:2104.08835}.

\bibitem[{Yoran et~al.(2021)Yoran, Talmor, and Berant}]{yoran2021turning}
Ori Yoran, Alon Talmor, and Jonathan Berant. 2021.
\newblock Turning tables: Generating examples from semi-structured tables for
  endowing language models with reasoning skills.
\newblock \emph{arXiv preprint arXiv:2107.07261}.

\bibitem[{Yu et~al.(2018)Yu, Zhang, Yang, Yasunaga, Wang, Li, Ma, Li, Yao,
  Roman, Zhang, and Radev}]{yu-etal-2018-spider}
Tao Yu, Rui Zhang, Kai Yang, Michihiro Yasunaga, Dongxu Wang, Zifan Li, James
  Ma, Irene Li, Qingning Yao, Shanelle Roman, Zilin Zhang, and Dragomir Radev.
  2018.
\newblock {S}pider: A large-scale human-labeled dataset for complex and
  cross-domain semantic parsing and text-to-{SQL} task.
\newblock In \emph{EMNLP}.

\bibitem[{Zellers et~al.(2018)Zellers, Bisk, Schwartz, and
  Choi}]{zellers-etal-2018-swag}
Rowan Zellers, Yonatan Bisk, Roy Schwartz, and Yejin Choi. 2018.
\newblock {SWAG}: A large-scale adversarial dataset for grounded commonsense
  inference.
\newblock In \emph{EMNLP}.

\bibitem[{Zellers et~al.(2019)Zellers, Holtzman, Bisk, Farhadi, and
  Choi}]{zellers-etal-2019-hellaswag}
Rowan Zellers, Ari Holtzman, Yonatan Bisk, Ali Farhadi, and Yejin Choi. 2019.
\newblock {H}ella{S}wag: Can a machine really finish your sentence?
\newblock In \emph{ACL}.

\bibitem[{Zhang et~al.(2018)Zhang, Liu, Liu, Gao, Duh, and
  Durme}]{Zhang2018ReCoRDBT}
Sheng Zhang, X.~Liu, J.~Liu, Jianfeng Gao, Kevin Duh, and Benjamin~Van Durme.
  2018.
\newblock Record: Bridging the gap between human and machine commonsense
  reading comprehension.
\newblock \emph{arXiv preprint arXiv:1810.12885}.

\bibitem[{Zhang et~al.(2015)Zhang, Zhao, and LeCun}]{zhang2015character}
Xiang Zhang, Junbo Zhao, and Yann LeCun. 2015.
\newblock Character-level convolutional networks for text classification.
\newblock In \emph{Advances in neural information processing systems}, pages
  649--657.

\bibitem[{Zhang et~al.(2019)Zhang, Baldridge, and He}]{zhang-etal-2019-paws}
Yuan Zhang, Jason Baldridge, and Luheng He. 2019.
\newblock {PAWS}: Paraphrase adversaries from word scrambling.
\newblock In \emph{NAACL-HLT}.

\bibitem[{Zhao et~al.(2021)Zhao, Wallace, Feng, Klein, and
  Singh}]{zhao2021calibrate}
Zihao Zhao, Eric Wallace, Shi Feng, Dan Klein, and Sameer Singh. 2021.
\newblock Calibrate before use: Improving few-shot performance of language
  models.
\newblock In \emph{International Conference on Machine Learning}, pages
  12697--12706. PMLR.

\bibitem[{Zhong et~al.(2021{\natexlab{a}})Zhong, Lee, Zhang, and
  Klein}]{zhong2021adapting}
Ruiqi Zhong, Kristy Lee, Zheng Zhang, and Dan Klein. 2021{\natexlab{a}}.
\newblock Adapting language models for zero-shot learning by meta-tuning on
  dataset and prompt collections.
\newblock \emph{arXiv preprint arXiv:2104.04670}.

\bibitem[{Zhong et~al.(2021{\natexlab{b}})Zhong, Lee, Zhang, and
  Klein}]{zhong2021meta}
Ruiqi Zhong, Kristy Lee, Zheng Zhang, and Dan Klein. 2021{\natexlab{b}}.
\newblock Meta-tuning language models to answer prompts better.
\newblock \emph{CoRR}.

\bibitem[{Zhong et~al.(2017)Zhong, Xiong, and Socher}]{zhongSeq2SQL2017}
Victor Zhong, Caiming Xiong, and Richard Socher. 2017.
\newblock Seq2sql: Generating structured queries from natural language using
  reinforcement learning.
\newblock \emph{arXiv preprint arXiv:1709.00103}.

\bibitem[{Zhou et~al.(2019)Zhou, Khashabi, Ning, and
  Roth}]{zhou-etal-2019-going}
Ben Zhou, Daniel Khashabi, Qiang Ning, and Dan Roth. 2019.
\newblock {``}going on a vacation{''} takes longer than {``}going for a
  walk{''}: A study of temporal commonsense understanding.
\newblock In \emph{EMNLP}.

\end{thebibliography}
\clearpage
\appendix

\section{Tables-to-tasks filtering}
\label{sec:tables_to_tasks_filtering}

Tab. \ref{tab:tables_filtering} shows the number of tables and tasks filtered at various stages of our tables-to-tasks procedure.

\begin{table}[h]
\resizebox{\columnwidth}{!}{
\begin{tabular}{|l r|}
\hline
tables initial & $50,820,216$ \\
rejected min rows & $-25,638,244$ \\
rejected non-english & $-23,034,542$ \\
\hline
tables remaining & $2,147,532$ \\
\hline
\hline
tasks initial & $5,646,614$ \\
rejected max domain & $-4,054,764$ \\
rejected min rows & $-99,226$ \\
rejected one-to-many & $-322,536$ \\
rejected min classes & $-157,199$ \\
rejected non-english output & $-561,622$ \\
rejected class balance & $-38,505$ \\
\hline
tasks remaining & $413,299$ \\
\hline

\end{tabular}}
\caption{Converting 50M tables into 400k tasks.}
\label{tab:tables_filtering}
\end{table}




\section{MetaICL experiment details}
\label{sec:metaicl_details}

This section provides training and evaluation details for our MetaICL experiments in $\S$\ref{sec:finetuning} and $\S$\ref{sec:distributions}. The datasets used in MetaICL train and test settings are taken from \textsc{CrossFit}~\citep{ye2021crossfit} and \textsc{UnifiedQA}~\citep{khashabi2020unifiedqa}, which in turn have been compiled from various other sources. The full list for all datasets and their citations are provided in Fig. \ref{tab:metaicl_splits}. We make use of 3 different task splits:

\paragraph{Test Tasks (52 tasks)} The union of all test tasks from the 7 task settings in \citet{min2021metaicl}.
\paragraph{Train Tasks (90 tasks)} Contains all tasks in \citet{min2021metaicl} except those which are Test Tasks. These tasks are only used as a source of NLP datasets in $\S$\ref{sec:distributions}.
\paragraph{Dev Tasks (50 tasks)} Contains all our Train Tasks except those which are not multiple-choice. These tasks are used for hyperparameter selection.

For hyperparameter selection, we finetune the GPT2-large model (774M)\footnote{GPT2-large LM \href{https://huggingface.co/gpt2-large}{https://huggingface.co/gpt2-large}} on \texttt{UnpredicTable-5k} and sweep over batch sizes $\{1, 8, 64\}$ and learning rates $\{5e^{-5}, 5e^{-6}, 5e^{-7}\}$. We select batch size = 1 and learning rate = $5e^{-6}$ based on Dev scores and use this for all MetaICL experiments. We train for 5 epochs and evaluate after each epoch, selecting the checkpoint with the highest mean Dev Tasks score. We report scores of the selected checkpoint evaluated on the Test Tasks. Each training and inference run is done on a single RTX8000 GPU. The duration of training varies by dataset size (training 5 epochs on \texttt{UnpredicTable-5k} takes $\sim$24 hours).

\section{Do Other Learning Algorithms Benefit from Table Data?}
\label{sec:additiona_setups}

Our main experiments use the MetaICL algorithm and benchmarks for training and evaluation. To understand how well our findings hold in other settings, we report additional experiments comparing \texttt{UnpredicTable-5k} against NLP datasets using different models, multi-task learning algorithms, and evaluation settings.

\subsection{CrossFit}

\citet{ye2021crossfit} introduce the Few-Shot Gym, a collection of 160 NLP tasks, and a problem setup called CrossFit. We focus on the \textit{Random} task partition of CrossFit where $\mathcal{D_\text{train}}$ and $\mathcal{D_\text{test}}$ contain 120 and 20 tasks respectively, sampled IID from the Few-Shot Gym.
For our learning algorithm, we adopt the best-performing method in \citet{ye2021crossfit}, MTL, which finetunes on $\mathcal{D_\text{train}}$ followed by finetuning on the few-shot training examples from a given target task in $\mathcal{D_\text{test}}$ (finetuning a separate model for each target task in $\mathcal{D_\text{test}}$).
We compare three different methods: MTL with $\mathcal{D_\text{train}}$ from the Few-Shot Gym, MTL with \texttt{UnpredicTable-5k} as $\mathcal{D_\text{train}}$, and Direct Finetuning (DF) which is a baseline without finetuning on any $\mathcal{D_\text{train}}$.
All experiments finetune a BART-Base \cite{lewis2019bart}, a pretrained encoder-decoder transformer model~\cite{vaswani2017attention}.


\begin{table}[h]
\centering
\begin{tabular}{|l|r|r|r|}
\hline
\textbf{Task} & \multicolumn{1}{l|}{\textbf{DF}} & \multicolumn{1}{l|}{\textbf{MTL}} & \multicolumn{1}{l|}{\textbf{Ours}} \\ \hline
glue-cola & 0.0 & 1.0 & 0.0 \\ \hline
crawl\_domain & 30.6 & 25.6 & 29.5 \\ \hline
ag\_news & 86.1 & 82.6 & 84.9 \\ \hline
ai2\_arc & 16.1 & 25.4 & 15.7 \\ \hline
wiki\_split & 79.6 & 80.0 & 78.4 \\ \hline
amazon\_polarity & 79.4 & 92.1 & 90.8 \\ \hline
blimp-...\_present & 99.4 & 98.5 & 97.8 \\ \hline
tweet\_eval-irony & 55.0 & 56.4 & 52.5 \\ \hline
ethos-disability & 75.8 & 77.7 & 71.3 \\ \hline
sglue-rte & 49.5 & 56.2 & 49.9 \\ \hline
circa & 46.3 & 44.8 & 48.3 \\ \hline
ethos-sexual\_orient. & 57.7 & 69.9 & 60.9 \\ \hline
hatexplain & 42.0 & 45.5 & 41.0 \\ \hline
race-high & 16.5 & 32.4 & 14.2 \\ \hline
glue-qnli & 60.5 & 74.2 & 56.9 \\ \hline
quoref & 24.7 & 41.8 & 23.3 \\ \hline
blimp-...npi\_scope & 70.9 & 97.1 & 82.6 \\ \hline
break-QDMR & 2.3 & 4.8 & 1.7 \\ \hline
yelp\_polarity & 40.6 & 93.5 & 56.2 \\ \hline
freebase-qa & 0.5 & 1.2 & 0.4 \\ \hline
\textbf{mean} & \textbf{46.7} & \textbf{49.1} & \textbf{47.8} \\ \hline
\end{tabular}
\caption{Results on the CrossFit benchmark. We compare the Direct Finetuning \textbf{DF} baseline (no multi-task learning) against multi-task learning on the NLP Few-shot Gym dataset (\textbf{MTL}) and multi-task learning with \texttt{UnpredicTable-5k} (\textbf{Ours}).}
\label{tab:crossfit}
\end{table}

\paragraph{Results} Tab. \ref{tab:crossfit} shows the full results. Compared to DF, MTL with our dataset improves results by a mean of +1.1\%. 3 out of 20 tasks improve by more than +10\% including \texttt{amazon\_polarity} and \texttt{yelp\_polarity}, which are also among the tasks with the largest improvements in MetaICL. MTL with \texttt{UnpredicTable-5k} is less helpful than MTL with curated NLP datasets (+2.4\% relative to DF), but still recovers 46\% of the relative improvement from finetuning on 120 curated NLP tasks.
Our results show that finetuning on UnpredicTable helps even with MTL (a different learning algorithm) on BART (a different LM). We see large gains on similar tasks as in MetaICL, which suggests that our data helps consistently on these tasks (and the observed gains are not just an artifact of MetaICL training).

\subsection{FLEX}

FLEX \cite{bragg2021flex} is a FSL benchmark that provides 11 NLP training tasks and 20 NLP test tasks, carefully chosen to evaluate various task transfer settings.
The baseline model is \textbf{$\text{UniFew}$}, which uses a UnifiedQA model \cite{khashabi2020unifiedqa} with a prompt that converts task examples into a multiple-choice question-answer format. The primary FLEX model is \textbf{$\text{UniFew}_\text{Meta}$}, which is $\text{UniFew}$ finetuned with the 11 FLEX training tasks. As in MetaICL, $\text{UniFew}_\text{Meta}$ finetuning uses $k$ examples in the input to maximize $\log P(y_{k+1} | x_1, y_1, \dots, x_k, y_k, x_{k+1})$.
Our approach (\textbf{Ours}) uses the same setup as $\text{UniFew}_\text{Meta}$ but replaces the FLEX training tasks with \texttt{UnpredicTable-5k}.
Evaluation for all models is done with FSL on the FLEX test tasks.

\begin{table}[h]
\centering
\begin{tabular}{|l|r|r|r|}
\hline
\textbf{Task} & \multicolumn{1}{l|}{\textbf{UniFew}} & \multicolumn{1}{l|}{\textbf{Ours}} & \multicolumn{1}{l|}{\textbf{$\text{UniFew}_\text{Meta}$}} \\ \hline
FewRel & 79.2 & 79.4 & 87.2 \\ \hline
HuffPost & 62.8 & 63.1 & 68.0 \\ \hline
Amazon & 79.5 & 79.4 & 82.1 \\ \hline
20News & 63.1 & 63.4 & 67.3 \\ \hline
Reuters & 94.5 & 95.5 & 96.3 \\ \hline
\hline
MR & 78.6 & 83.1 & 89.4 \\ \hline
CR & 90.1 & 92.0 & 93.3 \\ \hline
SNLI & 55.8 & 56.5 & 80.9 \\ \hline
SciTail & 64.9 & 65.5 & 83.6 \\ \hline
\hline
SUBJ & 60.5 & 63.7 & 68.7 \\ \hline
TREC & 58.1 & 62.9 & 60.0 \\ \hline
CoNLL & 44.3 & 44.0 & 58.6 \\ \hline
\hline
\textbf{Mean} & \textbf{69.3} & \textbf{70.7} & \textbf{77.9} \\ \hline
\end{tabular}
\caption{Results on the FLEX benchmark. We compare the pretraining-only \textbf{UniFew} model against the same model finetuned on the FLEX dataset (\textbf{Unifew-Meta}) and \texttt{UnpredicTable-5k} (\textbf{Ours}).}
\label{tab:flex}
\end{table}

\paragraph{Results}
Tab. \ref{tab:flex} shows our results. 
Training on our dataset improves over $\text{UniFew}$ for 10/12 tasks (mean +1.4\%, max +5.5\%). However, we do not approach the level of $\text{UniFew}_\text{Meta}$ (mean improvement +8.6\%). This discrepancy is likely because the FLEX training and test tasks have been chosen with overlapping domains/task types to study various transfer learning settings (see \citet{bragg2021flex} for details).
Nevertheless, the results show that our table tasks still lead to improvements in FLEX with a different model and test tasks.



\section{Clustering}
\label{sec:clustering_details}
Here, we describe the clustering procedure used to group \texttt{UnpredicTable-unique} tasks into narrow data subsets based on content.
For all examples in all tasks, we concatenate each $(x, y)$ example and obtain their embeddings from a pretrained GPT-2 model\footnote{The \texttt{stanford-crfm/eowyn-gpt2-medium-x777} model via the HuggingFace Transformers library.}. We average the resulting 1024-dimensional embeddings at a task level. We normalize each task embedding and apply a two-stage dimensionality reduction consisting of a PCA transformation to 128 dimensions followed by further reduction using UMAP (\citet{mcinnes2018umap-software}, $n_{\text{neighbors}} = 4$, $\text{d}_{\text{min}} = 0.0$) to 32 dimensions. We cluster the 32D task embeddings using the HDBSCAN algorithm \cite{mcinnes2017hdbscan} with a minimum cluster size of 60 and 400 minimum samples. This setup results in 30 task clusters plus an additional cluster (\texttt{cluster -1}) containing tasks that HDBSCAN rejected as noise. The cluster sizes range from $T=61$ to $T=5700$. We tested several hyperparameters for our clustering pipeline until we arrived at a setup with reasonable in-cluster content similarity (manual inspection).

\section{Task Quality Annotation Instructions}
\label{sec:task_quality_annotations}
Below, we display a condensed version of the instructions given to annotators for annotating the dataset into different task quality levels. The full instructions are available online\footnote{Full instructions for task quality annotations: \url{https://bit.ly/3veIWf7}\label{annoinsturl}}.


\paragraph{Introduction}
Thank you for agreeing to contribute annotations to our dataset! Here are some brief instructions to help you successfully complete this work.

\paragraph{Context}
We have a large number of \textbf{Tasks} created for training language models to learn a variety of skills. A standard example of a task is shown in Tab. \ref{tab:annotations_example_tasks} as Task 1.
This example closely resembles the Question-Answer form that is commonly encountered in human competency tests, but this is not the only valid form. More generally, a \textbf{Task} is simply a set of \textbf{input-output} pairs where the inputs map to outputs in a common and (given knowledge of the mapping) predictable way; given an input, an individual skilled in this task should be able to respond with the correct output. Another example of a valid task is shown in Tab. \ref{tab:annotations_example_tasks} as Task 2. In this case, the inputs are a set of issues that a user might be having, and the outputs suggest actions to address each issue.

\begin{centering}
\begin{table}[ht!]
\begin{tabular}[t]{|c|p{0.78\linewidth}|}
\hline
\multicolumn{2}{|c|}{\textbf{\textit{Examples of Tasks for Annotation}}} \\
\hline
\multicolumn{2}{|c|}{\textbf{Task 1}} \\
\hline
input & [Question] The parotid glands are located: [Answer] \\
output & cheek \\ [5pt]
\hdashline
input & [Question] The roof of the mouth is called the: [Answer] \\
output & hard palte \\ [5pt]
\hdashline
input & [Question] The bone that forms the posterior portion of the skull is the [Answer] \\
output & occipital bone \\ [5pt]
\hdashline
input & [Question] The lower jawbone is the [Answer] \\
output & mandible \\ [5pt]
\hline
\multicolumn{2}{|c|}{\textbf{Task 2}} \\
\hline
input & [If you want to ...] Get a page or site removed from Google [Then ...] \\
output & Submit a URL removal request. \\ [5pt]
\hdashline
input & [If you want to ...] Report spam [Then ...]  \\
output & Submit a spam report. \\ [5pt]
\hdashline
input & [If you want to ...] Report a copyright violation or the misuse of your content [Then ...] \\
output & File a DMCA takedown request. \\ [5pt]
\hdashline
input & [If you want to ...] Tell Google to crawl your site more slowly [Then ...] \\
output & Request a change in crawl rate. \\ [5pt]
\hdashline
input & [If you want to ...] Tell Google that your content is mistakenly being filtered by SafeSearch [Then ...] \\
output & Submit a SafeSearch issue. \\ [5pt]
\hline
\end{tabular}
\caption{Example tasks provided with the instructions for the task-quality annotation}
\label{tab:annotations_example_tasks}
\end{table}
\end{centering}

\paragraph{The Problem}
Our pool of tasks has been curated in an automated way from natural internet content, so they vary greatly in quality and form. It would be valuable to label each task’s quality so that we may investigate (1) what is the overall quality in our pool of tasks, and (2) how task quality affects the ability of language models to learn from it.

\paragraph{The Work}
In this session, you will classify a number of tasks in terms of how feasible and useful they are. Each task should be rated from 0-2, where 0 is “This task is not valid or useful at all” and 2 is “This task demonstrates an interesting and useful skill”.


\paragraph{Criteria of Class 0 (low rating) Tasks}

\begin{itemize}
    \item The input-output mapping appears nonsensical and/or arbitrary.
    \item The task is not in English.
    \item Would never be useful in any realistic setting / practicing this task does not build any generally-useful skills.
    \item Tests highly obscure knowledge that is not correlated with the input text (highly context-dependent knowledge, entertainment trivia on fan sites, product specifications, …)
    \item You would not even be able to tell if all output labels have been shuffled.
\end{itemize}

\paragraph{Criteria of Class 1 (medium rating) Tasks}
\begin{itemize}
    \item This class is a catch-all for tasks that are neither squarely Class 0 nor Class 2.
    \item The task is quite interesting, but its current form contains flaws that make it confusing or lacks enough context to do a good job of the task.
    \item You could narrow the space of possible options and guess the right answer with better-than-random accuracy (especially with the help of multiple-choice options).
    \item The task makes sense but is trivial or not interesting enough to be Class 2. For example, the output is just a copy of the input.
\end{itemize}

\paragraph{Criteria of Class 2 (high rating) Tasks}
\begin{itemize}
    \item The task is well-posed with enough context that an expert could give a reasonably correct answer most of the time.
    \item Demonstrates a skill that is definitely useful for real-world tasks, i.e. might be tested in an exam or competency test, or part of a job.
    \item Resembles the type of skill that is tested in typical NLP datasets. See "Examples from real NLP datasets" section in the full instructions\textsuperscript{\ref{annoinsturl}}.
\end{itemize}

\paragraph{Further notes}
\begin{itemize}
    \item These criteria are not a complete set of rules for membership, so based on the above you may make your own judgement regarding a new task that does not perfectly fit any criteria.
    \item We expect that the majority of our tasks will fall into either Class 0 or Class 1; fewer than 20\% of the tasks will meet the standard for Class 2.
    \item A single input may not always be enough to know what the task expects in the output; this is acceptable (even for Class 2) as long as the input-output mapping is clear after observing several demonstration pairs.
    \item The "Examples from real NLP datasets" section in the full instructions\textsuperscript{\ref{annoinsturl}} show the kinds of interesting tasks we would like to see in Class 2, but we expect (and encourage) that our tasks will span a wider variety that are still interesting and valuable.
\end{itemize}

\section{Examples of tasks}
\label{sec:task_examples}
In the following pages, we provide examples from various datasets discussed in the text:

\begin{enumerate}
    \item Quality-annotated (High)
    \item Quality-annotated (Med)
    \item Quality-annotated (Low)
    \item Single-website (support.google.com)
    \item Single-website (w3.org)
    \item Single-website (mmo-champion)
    \item Single-website (studystack.com)
    \item Cluster 7
    \item Cluster 8
    \item Cluster -1
    \item Cluster 3
    \item NLP train (2 best and 2 worst)
    \item NLP test (10 most-improving)
\end{enumerate}

\begin{figure*}
    \centering
    \begin{tabular}{p{\linewidth}}
    \hline
    \textbf{Train Tasks (90 tasks)} \\ \hline
    \small{ade\_corpus\_v2-classification~\citep{GURULINGAPPA2012885},
        ade\_corpus\_v2-dosage~\citep{GURULINGAPPA2012885},
        art~\citep{bhagavatula2020abductive},
        biomrc~\citep{pappas-etal-2020-biomrc},
        blimp-anaphor\_number\_agreement~\citep{warstadt2019blimp},
        blimp-ellipsis\_n\_bar\_2~\citep{warstadt2019blimp},
        blimp-sentential\_negation\_npi\_licensor\_present~\citep{warstadt2019blimp},
        blimp-sentential\_negation\_npi\_scope~\citep{warstadt2019blimp},
        boolq~\citep{clark-etal-2019-boolq},
        circa~\citep{louis-etal-2020-id},
        crows\_pairs~\citep{nangia-etal-2020-crows},
        discovery~\citep{sileo-etal-2019-mining},
        emotion~\citep{saravia-etal-2018-carer},
        ethos-directed\_vs\_generalized~\citep{Mollas2020ETHOSAO},
        ethos-disability~\citep{Mollas2020ETHOSAO},
        ethos-gender~\citep{Mollas2020ETHOSAO},
        ethos-sexual\_orientation~\citep{Mollas2020ETHOSAO},
        freebase\_qa~\citep{jiang-etal-2019-freebaseqa},
        gigaword~\citep{napoles-etal-2012-annotated},
        glue-cola~\citep{warstadt-etal-2019-neural},
        glue-sst2~\citep{socher-etal-2013-recursive},
        google\_wellformed\_query~\citep{faruqui-das-2018-identifying},
        hate\_speech\_offensive~\citep{hateoffensive},
        hatexplain~\citep{mathew2020hatexplain},
        health\_fact~\citep{kotonya-toni-2020-explainable-automated},
        hotpot\_qa~\citep{yang-etal-2018-hotpotqa},
        imdb~\citep{maas-etal-2011-learning},
        kilt\_ay2~\citep{hoffart-etal-2011-robust},
        kilt\_fever~\citep{thorne-etal-2018-fever},
        kilt\_hotpotqa~\citep{yang-etal-2018-hotpotqa},
        kilt\_nq~\citep{kwiatkowski-etal-2019-natural},
        kilt\_trex~\citep{elsahar-etal-2018-rex},
        kilt\_zsre~\citep{levy-etal-2017-zero},
        lama-conceptnet~\citep{petroni-etal-2019-language,petroni2020how},
        lama-google\_re~\citep{petroni-etal-2019-language,petroni2020how},
        lama-squad~\citep{petroni-etal-2019-language,petroni2020how},
        lama-trex~\citep{petroni-etal-2019-language,petroni2020how},
        liar~\citep{wang-2017-liar},
        mc\_taco~\citep{zhou-etal-2019-going},
        numer\_sense~\citep{lin-etal-2020-birds},
        onestop\_english~\citep{vajjala-lucic-2018-onestopenglish},
        piqa~\citep{bisk2019piqa},
        proto\_qa~\citep{boratko-etal-2020-protoqa},
        qa\_srl~\citep{he-etal-2015-question},
        quoref~\citep{dasigi-etal-2019-quoref}12,
        race-high~\citep{lai-etal-2017-race},
        race-middle~\citep{lai-etal-2017-race},
        ropes~\citep{lin-etal-2019-reasoning},
        rotten\_tomatoes~\citep{pang-lee-2005-seeing},
        search\_qa~\citep{Dunn2017SearchQAAN},
        sms\_spam~\citep{sms_spam},
        social\_i\_qa~\citep{sap-etal-2019-social},
        spider~\citep{yu-etal-2018-spider},
        squad-no\_context~\citep{rajpurkar-etal-2016-squad},
        squad-with\_context~\citep{rajpurkar-etal-2016-squad},
        superglue-multirc~\citep{khashabi-etal-2018-looking},
        superglue-record~\citep{Zhang2018ReCoRDBT},
        superglue-rte~\begin{tabular}[c]{@{}l@{}}\citep{dagan2005pascal, bar2006second}\citep{giampiccolo2007third, bentivogli2009fifth}\end{tabular},
        superglue-wic~\citep{pilehvar-camacho-collados-2019-wic},
        superglue-wsc~\citep{levesque2012winograd},
        trec~\citep{li-roth-2002-learning,hovy-etal-2001-toward},
        trec-finegrained~\citep{li-roth-2002-learning,hovy-etal-2001-toward},
        tweet\_eval-emoji~\citep{barbieri-etal-2020-tweeteval},
        tweet\_eval-emotion~\citep{barbieri-etal-2020-tweeteval},
        tweet\_eval-irony~\citep{barbieri-etal-2020-tweeteval},
        tweet\_eval-offensive~\citep{barbieri-etal-2020-tweeteval},
        tweet\_eval-sentiment~\citep{barbieri-etal-2020-tweeteval},
        tweet\_eval-stance\_abortion~\citep{barbieri-etal-2020-tweeteval},
        tweet\_eval-stance\_climate~\citep{barbieri-etal-2020-tweeteval},
        tweet\_eval-stance\_hillary~\citep{barbieri-etal-2020-tweeteval},
        tweet\_qa~\citep{xiong-etal-2019-tweetqa},
        unifiedqa:boolq~\citep{clark-etal-2019-boolq},
        unifiedqa:commonsenseqa~\citep{talmor-etal-2019-commonsenseqa},
        unifiedqa:drop~\citep{dua-etal-2019-drop},
        unifiedqa:narrativeqa~\citep{kocisky-etal-2018-narrativeqa},
        unifiedqa:natural\_questions\_with\_dpr\_para,
        unifiedqa:newsqa~\citep{trischler-etal-2017-newsqa},
        unifiedqa:physical\_iqa~\citep{bisk2019piqa},
        unifiedqa:quoref~\citep{dasigi-etal-2019-quoref},
        unifiedqa:race\_string~\citep{lai-etal-2017-race},
        unifiedqa:ropes~\citep{lin-etal-2019-reasoning},
        unifiedqa:social\_iqa~\citep{sap2019socialiqa},
        unifiedqa:squad1\_1~\citep{rajpurkar-etal-2016-squad},
        unifiedqa:squad2~\citep{rajpurkar-etal-2018-know},
        unifiedqa:winogrande\_xl~\citep{sakaguchi2019winogrande},
        web\_questions~\citep{berant-etal-2013-semantic},
        wikisql~\citep{zhongSeq2SQL2017},
        xsum~\citep{narayan-etal-2018-dont},
        yahoo\_answers\_topics~\href{https://webscope.sandbox.yahoo.com/catalog.php?datatype=l}{(link)},
        yelp\_review\_full~\citep{zhang2015character}} \\ \hline
    
    \textbf{Test Tasks (52 tasks)} \\ \hline
    \small{ag\_news~\href{http://groups.di.unipi.it/~gulli/AG_corpus_of_news_articles.html}{Gulli (link)},
        ai2\_arc~\citep{Clark2018ThinkYH},
        amazon\_polarity~\citep{McAuley2013HiddenFA},
        anli~\citep{nie-etal-2020-adversarial},
        climate\_fever~\citep{Diggelmann2020CLIMATEFEVERAD},
        codah~\citep{chen-etal-2019-codah},
        commonsense\_qa~\citep{talmor-etal-2019-commonsenseqa},
        cosmos\_qa~\citep{huang-etal-2019-cosmos},
        dbpedia\_14~\citep{Lehmann2015DBpediaA},
        dream~\citep{sun-etal-2019-dream},
        emo~\citep{chatterjee-etal-2019-semeval},
        ethos-national\_origin~\citep{Mollas2020ETHOSAO},
        ethos-race~\citep{Mollas2020ETHOSAO},
        ethos-religion~\citep{Mollas2020ETHOSAO},
        financial\_phrasebank~\citep{financial-phrasebank},
        glue-mnli~\citep{williams-etal-2018-broad},
        glue-mrpc~\citep{dolan-brockett-2005-automatically},
        glue-qnli~\citep{rajpurkar-etal-2016-squad},
        glue-qqp~(\url{data.quora.com/First-Quora-Dataset-Release-Question-Pairs}),
        glue-rte~\begin{tabular}[c]{@{}l@{}}\citep{dagan2005pascal, bar2006second}\citep{giampiccolo2007third, bentivogli2009fifth}\end{tabular},
        glue-wnli~\citep{levesque2012winograd},
        hate\_speech18~\citep{gibert2018hate},
        hellaswag~\citep{zellers-etal-2019-hellaswag},
        medical\_questions\_pairs~\citep{medical-qqp},
        openbookqa~\citep{mihaylov-etal-2018-suit},
        paws~\citep{zhang-etal-2019-paws},
        poem\_sentiment~\citep{sheng-uthus-2020-investigating},
        qasc~\citep{Khot_Clark_Guerquin_Jansen_Sabharwal_2020},
        quail~\citep{Rogers_Kovaleva_Downey_Rumshisky_2020},
        quarel~\citep{Tafjord_Clark_Gardner_Yih_Sabharwal_2019},
        quartz-no\_knowledge~\citep{tafjord-etal-2019-quartz},
        quartz-with\_knowledge~\citep{tafjord-etal-2019-quartz},
        sciq~\citep{welbl-etal-2017-crowdsourcing},
        scitail~\citep{scitail},
        sick~\citep{marelli-etal-2014-sick},
        superglue-cb~\citep{Marneffe_Simons_Tonhauser_2019},
        superglue-copa~\citep{gordon-etal-2012-semeval},
        swag~\citep{zellers-etal-2018-swag},
        tab\_fact~\citep{Chen2020TabFact},
        tweet\_eval-hate~\citep{barbieri-etal-2020-tweeteval},
        tweet\_eval-stance\_atheism~\citep{barbieri-etal-2020-tweeteval},
        tweet\_eval-stance\_feminist~\citep{barbieri-etal-2020-tweeteval},
        unifiedqa:ai2\_science\_middle~(\url{data.allenai.org/ai2-science-questions}),
        unifiedqa:mctest~\citep{richardson-etal-2013-mctest},
        unifiedqa:openbookqa~\citep{mihaylov-etal-2018-suit},
        unifiedqa:openbookqa\_with\_ir,
        unifiedqa:qasc~\citep{khot2019qasc},
        unifiedqa:qasc\_with\_ir,
        wiki\_qa~\citep{yang-etal-2015-wikiqa},
        wino\_grande~\citep{Sakaguchi_Le_Bras_Bhagavatula_Choi_2020},
        wiqa~\citep{tandon-etal-2019-wiqa},
        yelp\_polarity~\citep{zhang2015character}} \\ \hline
    
    \textbf{Dev Tasks (50 tasks)} \\ \hline
    \small{ade\_corpus\_v2-classification,
        art,
        biomrc,
        blimp-anaphor\_number\_agreement,
        blimp-ellipsis\_n\_bar\_2,
        blimp-sentential\_negation\_npi\_licensor\_present,
        blimp-sentential\_negation\_npi\_scope,
        boolq,
        circa,
        crows\_pairs,
        discovery,
        emotion,
        ethos-directed\_vs\_generalized,
        ethos-disability,
        ethos-gender,
        ethos-sexual\_orientation,
        glue-cola,
        glue-sst2,
        google\_wellformed\_query,
        hate\_speech\_offensive,
        hatexplain,
        health\_fact,
        imdb,
        kilt\_fever,
        liar,
        mc\_taco,
        numer\_sense,
        onestop\_english,
        piqa,
        race-high,
        race-middle,
        rotten\_tomatoes,
        sms\_spam,
        social\_i\_qa,
        superglue-multirc,
        superglue-rte,
        superglue-wic,
        superglue-wsc,
        trec,
        trec-finegrained,
        tweet\_eval-emoji,
        tweet\_eval-emotion,
        tweet\_eval-irony,
        tweet\_eval-offensive,
        tweet\_eval-sentiment,
        tweet\_eval-stance\_abortion,
        tweet\_eval-stance\_climate,
        tweet\_eval-stance\_hillary,
        yahoo\_answers\_topics,
        yelp\_review\_full} \\ \hline
    \end{tabular}
    \caption{All the task datasets used in our MetaICL experiments, along with citations of their original source. Dev Tasks are a subset of Train Tasks so citations are not repeated.}
    \label{tab:metaicl_splits}
\end{figure*}

\newpage
\begin{figure*}[ht!]
\begin{tabular}[ht!]{|c|p{0.9\linewidth}|}
\hline
\multicolumn{2}{|c|}{\textbf{\textit{quality\_annotated : High}}} \\
\hline
\multicolumn{2}{|c|}{\textbf{Task 1} (6 examples)} \\
\hline
input & [Format option] Heading 3 [What it will look like]   \\
output & is a sub-header and can be used as a sub-section heading  \\ [5pt]
\hdashline
input & [Format option] Code / preformatted [What it will look like]   \\
output & Technical text that should be displayed in a fixed-width font  \\ [5pt]
\hdashline
input & [Format option] Heading 5 [What it will look like]   \\
output & is the smallest sub-header option  \\ [5pt]
\hline
\multicolumn{2}{|c|}{\textbf{Task 2} (10 examples)} \\
\hline
input & [No.] 07 [Answer] Sahara desert [Question]   \\
output & The biggest desert in the world is the  \\ [5pt]
\hdashline
input & [No.] 02 [Answer] Nile [Question]   \\
output & The longest river in the world is the  \\ [5pt]
\hdashline
input & [No.] 05 [Answer] Everest [Question]   \\
output & The highest mountain in the world is the  \\ [5pt]
\hline
\multicolumn{2}{|c|}{\textbf{Task 3} (6 examples)} \\
\hline
input & [property] monitorType [applies to] all [description] one of counter, guage, string [type] \\
output & enum \\ [5pt]
\hdashline
input & [property] observedAttribute [applies to] all [description] the attribute being observed [type] \\
output & string \\ [5pt]
\hdashline
input & [property] initThreshold [applies to] counter [description] initial threshold value [type] \\
output & number \\ [5pt]
\hline
\multicolumn{2}{|c|}{\textbf{Task 4} (14 examples)} \\
\hline
input & [Verse] 14 [King James Version] And she lay at his feet until the morning: and she rose up before one could know another. And he said, Let it not be known that a woman came into the floor. So she lay at his feet until morning. She got up before either could know the other. He said, "Don't let it be known that a woman came into the threshing-floor." [Analysis] \\
output & Boaz wants to avoid scandal. \\ [5pt]
\hdashline
input & [Verse] 5 [King James Version] And she said unto her, All that thou sayest unto me I will do. Ruth said to her, "I will do everything you say." [Analysis] \\
output & What Ruth must have thought of these orders, none can speculate. \\ [5pt]
\hdashline
input & [Verse] 1 [King James Version] Then Naomi her mother in law said unto her, My daughter, shall I not seek rest for thee, that it may be well with thee? Now Naomi, mother-in-law of Ruth, said to her, "My daughter, I should find you a place of rest, that will be good for you. [Analysis] \\
output & Naomi wants to settle Ruth properly. \\ [5pt]
\hline
\end{tabular}
\end{figure*}

\newpage
\begin{figure*}[ht!]
\begin{tabular}[ht!]{|c|p{0.9\linewidth}|}
\hline
\multicolumn{2}{|c|}{\textbf{\textit{quality\_annotated : Med}}} \\
\hline
\multicolumn{2}{|c|}{\textbf{Task 1} (11 examples)} \\
\hline
input & [Symptom] Sore Throat [Cold] Sore throat is commonly present with a cold. [Flu] Sore throat is not commonly present with the flu. [Allergies]   \\
output & Sore throat is sometimes present if enough post-nasal drainage occurs.  \\ [5pt]
\hdashline
input & [Symptom] Sudden Symptoms [Cold] Cold symptoms tend to develop over a few days. [Flu] The flu has a rapid onset within 3-6 hours. The flu hits hard and includes sudden symptoms like high fever, aches and pains. [Allergies]   \\
output & Rapid onset.  \\ [5pt]
\hdashline
input & [Symptom] Aches [Cold] Slight body aches and pains can be part of a cold. [Flu] Severe aches and pains are common with the flu. [Allergies]   \\
output & No aches and pains.  \\ [5pt]
\hline
\multicolumn{2}{|c|}{\textbf{Task 2} (9 examples)} \\
\hline
input & [0] Space Requirements Larger due to the existence of aggregation structures and history data; requires more indexes than OLTP  \\
output & Can be relatively small if historical data is archived  \\ [5pt]
\hdashline
input & [0] Backup and Recovery Instead of regular backups, some environments may consider simply reloading the OLTP data as a recovery method  \\
output & Backup religiously; operational data is critical to run the business, data loss is likely to entail significant monetary loss and legal liability  \\ [5pt]
\hdashline
input & [0] Queries Often complex queries involving aggregations  \\
output & Relatively standardized and simple queries Returning relatively few records  \\ [5pt]
\hline
\multicolumn{2}{|c|}{\textbf{Task 3} (7 examples)} \\
\hline
input & [Action] Add a point to an editable shape [Shortcut]   \\
output & Option-click the shape edge where you want to add a point  \\ [5pt]
\hdashline
input & [Action] Change a curved point of an editable shape into a corner point [Shortcut]   \\
output & Double-click the curved point  \\ [5pt]
\hdashline
input & [Action] Delete a point of an editable shape [Shortcut]   \\
output & Click point and press Delete  \\ [5pt]
\hline
\multicolumn{2}{|c|}{\textbf{Task 4} (8 examples)} \\
\hline
input & [0] Length [1] meter [2]  \\
output & distance light travels in a vacuum \\ [5pt]
\hdashline
input & [0] Time [1] second [2]  \\
output & oscillations of the cesium atom \\ [5pt]
\hdashline
input & [0] Electric current [1] ampere [2]  \\
output & attraction between two wires \\ [5pt]
\hline
\end{tabular}
\end{figure*}

\newpage
\begin{figure*}[ht!]
\begin{tabular}[ht!]{|c|p{0.9\linewidth}|}
\hline
\multicolumn{2}{|c|}{\textbf{\textit{quality\_annotated : Low}}} \\
\hline
\multicolumn{2}{|c|}{\textbf{Task 1} (285 examples)} \\
\hline
input & [Career Cluster] Manufacturing [Career Title] Stationary Engineers and Boiler Operators [Nontraditional for...]   \\
output & Women  \\ [5pt]
\hdashline
input & [Career Cluster] Health Science [Career Title] Health Care Social Workers [Nontraditional for...]   \\
output & Men  \\ [5pt]
\hdashline
input & [Career Cluster] Government and Public Administration [Career Title] Government Program Eligibility Interviewers [Nontraditional for...]   \\
output & Men  \\ [5pt]
\hline
\multicolumn{2}{|c|}{\textbf{Task 2} (8 examples)} \\
\hline
input & [RESTRICTED] YES CONFIDENTIAL [UNRESTRICTED]   \\
output & NO (Sensitive/need to know)  \\ [5pt]
\hdashline
input & [RESTRICTED] Available COUNSELING SERVICES [UNRESTRICTED]   \\
output & Available  \\ [5pt]
\hdashline
input & [RESTRICTED] Active Duty Military Only ELIGIBILITY [UNRESTRICTED]   \\
output & All personnel  \\ [5pt]
\hline
\multicolumn{2}{|c|}{\textbf{Task 3} (6 examples)} \\
\hline
input & [Talent Cards] Beat Back [Type]   \\
output & Melee  \\ [5pt]
\hdashline
input &  [Type]   \\
output & Insanity  \\ [5pt]
\hdashline
input & [Talent Cards] Clear Minded [Type]   \\
output & Focus  \\ [5pt]
\hline
\multicolumn{2}{|c|}{\textbf{Task 4} (10 examples)} \\
\hline
input & [Directive] odbc.default\_db [Master Value] no value [Local Value]   \\
output & no value  \\ [5pt]
\hdashline
input & [Directive] odbc.defaultlrl [Master Value] return up to 4096 bytes [Local Value]   \\
output & return up to 4096 bytes  \\ [5pt]
\hdashline
input & [Directive] odbc.defaultbinmode [Master Value] return as is [Local Value]   \\
output & return as is  \\ [5pt]
\hline
\end{tabular}
\end{figure*}

\newpage
\begin{figure*}[ht!]
\begin{tabular}[ht!]{|c|p{0.9\linewidth}|}
\hline
\multicolumn{2}{|c|}{\textbf{\textit{single\_website\_tables : support.google.com}}} \\
\hline
\multicolumn{2}{|c|}{\textbf{Task 1} (6 examples)} \\
\hline
input & [If you want to ...] Report a copyright violation or the misuse of your content [Then ...]   \\
output & File a DMCA takedown request.  \\ [5pt]
\hdashline
input & [If you want to ...] Tell Google to crawl your site more slowly [Then ...]   \\
output & Request a change in crawl rate.  \\ [5pt]
\hdashline
input & [If you want to ...] Get a site added back to Google [Then ...]   \\
output & If your site was distributing malware, and is now clean, request a malware review. If your site was showing spam, but is now clean, submit a reconsideration request. If your site was in violation of the Webmaster Guidelines, but is now clean, submit  \textit{{... (Truncated)}} \\ [5pt]
\hline
\multicolumn{2}{|c|}{\textbf{Task 2} (6 examples)} \\
\hline
input & [Term] Impressions [Search Console usage] Used exclusively for Google Search impressions [Analytics usage]   \\
output & Used for both AdWords impressions and Google Search impressions  \\ [5pt]
\hdashline
input & [Term] CTR [Search Console usage] Clickthrough rate. Clicks/Impressions for Google Search clicks. [Analytics usage]   \\
output & Clickthrough rate. Clicks/Impressions for both AdWords and Google Search clicks.  \\ [5pt]
\hdashline
input & [Term] Average Position [Search Console usage] Average ranking in Google Search results [Analytics usage]   \\
output & Average ranking in Google Search results  \\ [5pt]
\hline
\multicolumn{2}{|c|}{\textbf{Task 3} (7 examples)} \\
\hline
input & [Setting] Devices [Description] Campaigns target all types of devices, which include desktops, tablets, and mobile devices. Later, you can choose to customize ads for different devices. [Learn more]   \\
output & Types of mobile ads  \\ [5pt]
\hdashline
input & [Setting] Locations and languages [Description] Your campaign’s ads are eligible to show to customers in your targeted geographic locations, or to customers who have selected your targeted language as their interface language. We recommend choosing t \textit{{... (Truncated)}} \\
output & Location and language targeting  \\ [5pt]
\hdashline
input & [Setting] Type [Description] The campaign type determines which settings we'll show you as you create or edit your campaign. The type you choose tailors the campaign setup to just what's appropriate for your goals, eliminating unrelated features. We  \textit{{... (Truncated)}} \\
output & Choosing the campaign type that's right for you  \\ [5pt]
\hline
\multicolumn{2}{|c|}{\textbf{Task 4} (6 examples)} \\
\hline
input & [Then ...] File a DMCA takedown request. [If you want to ...]   \\
output & Report a copyright violation or the misuse of your content  \\ [5pt]
\hdashline
input & [Then ...] Submit a URL removal request. [If you want to ...]   \\
output & Get a page or site removed from Google  \\ [5pt]
\hdashline
input & [Then ...] If your site was distributing malware, and is now clean, request a malware review. If your site was showing spam, but is now clean, submit a reconsideration request. If your site was in violation of the Webmaster Guidelines, but is now cle \textit{{... (Truncated)}} \\
output & Get a site added back to Google  \\ [5pt]
\hline
\end{tabular}
\end{figure*}

\newpage
\begin{figure*}[ht!]
\begin{tabular}[ht!]{|c|p{0.9\linewidth}|}
\hline
\multicolumn{2}{|c|}{\textbf{\textit{single\_website\_tables : w3.org}}} \\
\hline
\multicolumn{2}{|c|}{\textbf{Task 1} (23 examples)} \\
\hline
input & [Keyword] week [Data type] A date consisting of a week-year number and a week number with no time zone [Control type] A week control [State]   \\
output & Week  \\ [5pt]
\hdashline
input & [Keyword] hidden [Data type] An arbitrary string [Control type] n/a [State]   \\
output & Hidden  \\ [5pt]
\hdashline
input & [Keyword] password [Data type] Text with no line breaks (sensitive information) [Control type] A text field that obscures data entry [State]   \\
output & Password  \\ [5pt]
\hline
\multicolumn{2}{|c|}{\textbf{Task 2} (6 examples)} \\
\hline
input & [Attribute Name] next [Details]   \\
output & an ECMAScript expression which returns the URI of the CCXML document to be fetched.  \\ [5pt]
\hdashline
input & [Attribute Name] timeout [Details]   \\
output & is an ECMAScript expression returning a string in CSS2 [CSS2] format interpreted as a time interval. The interval begins when the is executed. The fetch will fail if not completed at the end of this interval. A failed fetch will return the error.fetc \textit{{... (Truncated)}} \\ [5pt]
\hdashline
input & [Attribute Name] synch [Details]   \\
output & is an ECMAScript left-hand-side expression that is set to the fetch completion event. The specification of this attribute in a implies a blocking fetch, which will be executed synchronously. If this attribute is not specified, the fetch is asynchrono \textit{{... (Truncated)}} \\ [5pt]
\hline
\multicolumn{2}{|c|}{\textbf{Task 3} (7 examples)} \\
\hline
input & [Function] DeleteScope [Arguments] name(optional) [Description] Removes a scope from the scope stack. If no name is provided, the topmost scope is removed. Otherwise the scope with provided name is removed. A Failure status is returned if the stack i \textit{{... (Truncated)}} \\
output & Success or Failure  \\ [5pt]
\hdashline
input & [Function] CreateScope [Arguments] name(optional) [Description] Creates a new scope object and pushes it on top of the scope stack. If no name is provided the scope is anonymous and may be accessed only when it on the top of the scope stack. A Failur \textit{{... (Truncated)}} \\
output & Success or Failure  \\ [5pt]
\hdashline
input & [Function] UpdateVariable [Arguments] variableName, newValue, scopeName(optional) [Description] Assigns a new value to the variable specified. If scopeName is not specified, the variable is accessed in the topmost scope on the stack. A Failure status \textit{{... (Truncated)}} \\
output & Success or Failure  \\ [5pt]
\hline
\multicolumn{2}{|c|}{\textbf{Task 4} (9 examples)} \\
\hline
input & [Event Type] help [Action] reprompt [Audio Provided]   \\
output & yes  \\ [5pt]
\hdashline
input & [Event Type] noinput [Action] reprompt [Audio Provided]   \\
output & no  \\ [5pt]
\hdashline
input & [Event Type] exit [Action] exit interpreter [Audio Provided]   \\
output & no  \\ [5pt]
\hline
\end{tabular}
\end{figure*}

\newpage
\begin{figure*}[ht!]
\begin{tabular}[ht!]{|c|p{0.9\linewidth}|}
\hline
\multicolumn{2}{|c|}{\textbf{\textit{single\_website\_tables : mmo-champion.com}}} \\
\hline
\multicolumn{2}{|c|}{\textbf{Task 1} (15 examples)} \\
\hline
input & [Level] 384 [Type] Leather [Spec] Feral [Slot] Legs [Name]   \\
output & Deep Earth Legguards  \\ [5pt]
\hdashline
input & [Level] 384 [Type] Leather [Spec] Feral [Slot] Chest [Name]   \\
output & Deep Earth Raiment  \\ [5pt]
\hdashline
input & [Level] 384 [Type] Leather [Spec] Restoration [Slot] Shoulder [Name]   \\
output & Deep Earth Mantle  \\ [5pt]
\hline
\multicolumn{2}{|c|}{\textbf{Task 2} (23 examples)} \\
\hline
input & [Level] 384 [Type] Tier 13 [Slot] Token [Name] Crown of the Corrupted Protector [Instance] Dragon Soul [Boss] LFR Warmaster Blackhorn [Spec]   \\
output & Armor  \\ [5pt]
\hdashline
input & [Level] 384 [Type] Trinket [Slot] Trinket [Name] Bone-Link Fetish [Instance] Dragon Soul [Boss] LFR All Bosses Except Deathwing [Spec]   \\
output & Melee  \\ [5pt]
\hdashline
input & [Level] 384 [Type] Mace [Slot] Two-Hand [Name] Ataraxis, Cudgel of the Warmaster [Instance] Dragon Soul [Boss] LFR Warmaster Blackhorn [Spec]   \\
output & Melee  \\ [5pt]
\hline
\multicolumn{2}{|c|}{\textbf{Task 3} (12 examples)} \\
\hline
input & [ilvl] 85 [Type] Enchant [Item] Lesser Inscription of Charged Lodestone [Slot]   \\
output & Shoulder  \\ [5pt]
\hdashline
input & [ilvl] 346 [Type] Finger [Spec] Physical DPS [Item] Terrath's Signet of Balance [Slot]   \\
output & Finger  \\ [5pt]
\hdashline
input & [ilvl] 346 [Type] Finger [Spec] Melee [Item] Gorsik's Band of Shattering [Slot]   \\
output & Finger  \\ [5pt]
\hline
\multicolumn{2}{|c|}{\textbf{Task 4} (77 examples)} \\
\hline
input & [Level] 522 [Type] Mail [Spec] Physical DPS [Slot] Chest [Name] Carapace of Segmented Scale [Req. Standing]   \\
output & Revered  \\ [5pt]
\hdashline
input & [Level] 522 [Type] Leather [Spec] Physical DPS [Slot] Waist [Name] Darkfang Belt [Req. Standing]   \\
output & Revered  \\ [5pt]
\hdashline
input & [Level] 522 [Type] Trinket [Slot] Trinket [Name] Steadfast Talisman of the Shado-Pan Assault [Req. Standing]   \\
output & Friendly  \\ [5pt]
\hline
\end{tabular}
\end{figure*}

\newpage
\begin{figure*}[ht!]
\begin{tabular}[ht!]{|c|p{0.9\linewidth}|}
\hline
\multicolumn{2}{|c|}{\textbf{\textit{single\_website\_tables : studystack.com}}} \\
\hline
\multicolumn{2}{|c|}{\textbf{Task 1} (24 examples)} \\
\hline
input & [Answer] hard palte [Question]   \\
output & The roof of the mouth is called the:  \\ [5pt]
\hdashline
input & [Answer] middle ear [Question]   \\
output & The malleus, incus, and stapes are located in the:  \\ [5pt]
\hdashline
input & [Answer] Volar [Question]   \\
output & The palm of the hand is called what?  \\ [5pt]
\hline
\multicolumn{2}{|c|}{\textbf{Task 2} (15 examples)} \\
\hline
input & [Answer] Evert/eversion [Question]   \\
output & Turning outward, typically used to describe ankle motion.  \\ [5pt]
\hdashline
input & [Answer] Gliding motion [Question]   \\
output & Occurs when one bone slides over another. EX. kneecap  \\ [5pt]
\hdashline
input & [Answer] Invert/inversion [Question]   \\
output & Turning inward, typically used to describe ankle motion,  \\ [5pt]
\hline
\multicolumn{2}{|c|}{\textbf{Task 3} (13 examples)} \\
\hline
input & [Definition] freewriting, clustering, mapping, questioning, brainstorming [Term]   \\
output & prewriting techniques.  \\ [5pt]
\hdashline
input & [Definition] 5 senses, be specific, use comparisions, similes, metophores. Eliminate fluff words [Term]   \\
output & good writing techniques  \\ [5pt]
\hdashline
input & [Definition] (1) a topic and (2) a controlling idea [Term]   \\
output & Two parts of a topic sentence  \\ [5pt]
\hline
\multicolumn{2}{|c|}{\textbf{Task 4} (9 examples)} \\
\hline
input & [Definition] the amount of space something takes up [Term]   \\
output & Mass  \\ [5pt]
\hdashline
input & [Definition] a mixture made up of particles that are uniformly y distributed [Term]   \\
output & homogeneous mixture  \\ [5pt]
\hdashline
input & [Definition] the science of matter and how it changes [Term]   \\
output & Chemistry  \\ [5pt]
\hline
\end{tabular}
\end{figure*}

\newpage
\begin{figure*}[ht!]
\begin{tabular}[ht!]{|c|p{0.9\linewidth}|}
\hline
\multicolumn{2}{|c|}{\textbf{\textit{cluster\_tables : 7}}} \\
\hline
\multicolumn{2}{|c|}{\textbf{Task 1} (7 examples)} \\
\hline
input & [Cookie Name] \_\_utmb [Cookie Length] 30 minutes [Description]   \\
output & Establish and continue a user session on the site  \\ [5pt]
\hdashline
input & [Cookie Name] \_\_utmz [Cookie Length] 6 months [Description]   \\
output & Used to track traffic sources and page navigation  \\ [5pt]
\hdashline
input & [Cookie Name] \_UKWM [Cookie Length] 2 years [Description]   \\
output & Used to identify traffic sources  \\ [5pt]
\hline
\multicolumn{2}{|c|}{\textbf{Task 2} (8 examples)} \\
\hline
input & [Cookie Name or Service] MoodleSessionTest MoodleSession MoodleID\_ [Purpose]   \\
output & Our virtual learning environment, Moodle, uses cookies to record when visitors have successfully logged into the service.  \\ [5pt]
\hdashline
input & [Cookie Name or Service] ASPSESSIONIDCQBSDQCQ [Purpose]   \\
output & This is a functional cookie that does not contain any personal information and is automatically removed when the visitor closes their web browser.  \\ [5pt]
\hdashline
input & [Cookie Name or Service] CAKEPHP [Purpose]   \\
output & This is a functional cookie that does not contain any personal information and is automatically removed when the visitor closes their web browser.  \\ [5pt]
\hline
\multicolumn{2}{|c|}{\textbf{Task 3} (9 examples)} \\
\hline
input & [Cookie] guest\_id, ki [Information]   \\
output & These cookies allow you to access the Twitter feed on the homepage.  \\ [5pt]
\hdashline
input & [Cookie] use\_hitbox [Information]   \\
output & This is downloaded when you play an embedded YouTube video.  \\ [5pt]
\hdashline
input & [Cookie] BX, localization [Information]   \\
output & These cookies are downloaded by Flickr if you visit the page with the MEI Conference 2010 Photographs slideshow.  \\ [5pt]
\hline
\multicolumn{2}{|c|}{\textbf{Task 4} (12 examples)} \\
\hline
input & [Cookie] pmx\_cbtstat\{ID\} [Origin] www.whymsical.com [Persistency] Current session only [Information and Usage]   \\
output & These cookies are set to records the expand/collapse state for a CBT Navigator block content.  \\ [5pt]
\hdashline
input & [Cookie] pmx\_YOfs [Origin] www.whymsical.com [Persistency] Page load time [Information and Usage]   \\
output & This cookie will probably never see you. It is set on portal actions like click on a page number. The cookie is evaluated on load the desired page and then deleted. It is used to restore the vertical screen position as before the click.  \\ [5pt]
\hdashline
input & [Cookie] AWNUTSWhymsicalcom [Origin] www.whymsical.com [Persistency] Expires according to user-chosen session duration [Information and Usage]   \\
output & If you log-in as a member of this site, this cookie contains your user name, an encrypted hash of your password and the time you logged-in. It is used by the site software to ensure that features such as indicating new Forum and Private messages are  \textit{{... (Truncated)}} \\ [5pt]
\hline
\end{tabular}
\end{figure*}

\newpage
\begin{figure*}[ht!]
\begin{tabular}[ht!]{|c|p{0.9\linewidth}|}
\hline
\multicolumn{2}{|c|}{\textbf{\textit{cluster\_tables : 8}}} \\
\hline
\multicolumn{2}{|c|}{\textbf{Task 1} (7 examples)} \\
\hline
input & [0] Appearance [Scholarly Journals] Plain, “serious” cover Text with black \& white graphs, charts, and photographs which \textit{{... (Truncated)}} \\
output & Generally glossy cover Color photographs and illustrations used to support the article as well as draw in readers  \\ [5pt]
\hdashline
input & [0] Examples [Scholarly Journals] American Journal of Education Journal of the Evangelical Theological Society Modern Fiction Studies [Trade Journals]   \\
output & Indiana Business Instrumentalist Preaching  \\ [5pt]
\hdashline
input & [0] Validity [Scholarly Journals] Articles reviewed and evaluated by other experts in the field / discipline (peer reviewed / \textit{{... (Truncated)}} \\
output & Articles may be reviewed by one editor with knowledge related to the topic  \\ [5pt]
\hline
\multicolumn{2}{|c|}{\textbf{Task 2} (15 examples)} \\
\hline
input & [DATABASE TITLE] Engineered Materials Abstracts [FULL DESCRIPTION] Comprehensive index to world literature on engineered \textit{{... (Truncated)}} \\
output & no  \\ [5pt]
\hdashline
input & [DATABASE TITLE] Engineering Research Database [FULL DESCRIPTION] The ProQuest Engineering Research Database covers the  \textit{{... (Truncated)}} \\
output & no  \\ [5pt]
\hdashline
input & [DATABASE TITLE] ENGnetBASE [FULL DESCRIPTION] The ENGnetBase eBook collection includes over 2300 cutting-edge and bestselling  \textit{{... (Truncated)}} \\
output & yes  \\ [5pt]
\hline
\multicolumn{2}{|c|}{\textbf{Task 3} (20 examples)} \\
\hline
input & [Access] Website [2] Choose My Plate The new food and dietary guidelines! Also included are related links such as: farmer's markets, nutrition labels and food safety. Created by the USDA. [Subject]   \\
output & Health \& Nutrition  \\ [5pt]
\hdashline
input & [Access] Website [2] Library of Congress; Performing Arts Encyclopedia This is an amzing guide to the performing arts. You can \textit{{... (Truncated)}} \\
output & Art  \\ [5pt]
\hdashline
input & [Access] Library Card Required [2] Encyclopedia Britannica This encyclopedia has A LOT of information, which is great, but  \textit{{... (Truncated)}} \\
output & Cultures  \\ [5pt]
\hline
\multicolumn{2}{|c|}{\textbf{Task 4} (6 examples)} \\
\hline
input & [Time Frame of Event] Seconds/minutes/hours Provides sketchy details, may be inaccurate but good for firsthand accounts [Information Resource]   \\
output & Television/radio/internet  \\ [5pt]
\hdashline
input & [Time Frame of Event] Six months or more In depth analysis of event written by experts in their field. In most cases,  \textit{{... (Truncated)}} \\
output & Scholarly Journals  \\ [5pt]
\hdashline
input & [Time Frame of Event] Next day or two More details and greater accuracy, the first rough draft of history [Information Resource]   \\
output & Newspapers  \\ [5pt]
\hline
\end{tabular}
\end{figure*}

\newpage
\begin{figure*}[ht!]
\begin{tabular}[ht!]{|c|p{0.9\linewidth}|}
\hline
\multicolumn{2}{|c|}{\textbf{\textit{cluster\_tables : -1}}} \\
\hline
\multicolumn{2}{|c|}{\textbf{Task 1} (7 examples)} \\
\hline
input & [Domain Name] TinyHomeForSale.com [Price] \$1,999 [Buy] Buy it Now [Keyword]   \\
output & Tiny Home For Sale  \\ [5pt]
\hdashline
input & [Domain Name] DomainSalesHistory.com [Price] Offer [Buy] Buy it Now [Keyword]   \\
output & Domain Sales History  \\ [5pt]
\hdashline
input & [Domain Name] NearbyForSale.com [Price] \$999 [Buy] Buy it Now [Keyword]   \\
output & Nearby For Sale  \\ [5pt]
\hline
\multicolumn{2}{|c|}{\textbf{Task 2} (8 examples)} \\
\hline
input & [You are...] Supportive [You should have...]   \\
output & A strong stomach  \\ [5pt]
\hdashline
input & [You are...] Dependable [You should have...]   \\
output & Good ethical standards  \\ [5pt]
\hdashline
input & [You are...] Organized [You should have...]   \\
output & Excellent attention to detail  \\ [5pt]
\hline
\multicolumn{2}{|c|}{\textbf{Task 3} (10 examples)} \\
\hline
input & [Indonesian] perangko [English]   \\
output & stamp  \\ [5pt]
\hdashline
input & [Indonesian] surat [English]   \\
output & letter  \\ [5pt]
\hdashline
input & [Indonesian] terdaftar [English]   \\
output & registered mail  \\ [5pt]
\hline
\multicolumn{2}{|c|}{\textbf{Task 4} (9 examples)} \\
\hline
input & [Endpoint/Outcome Measure] Vertebral Morphometry (6-point, 95-point) [Modality] X-Ray, DXA, CT [Description]   \\
output & Automatic identification of vertebral body margins  \\ [5pt]
\hdashline
input & [Endpoint/Outcome Measure] Microarchitecture [Modality] MRI, High resolution QCT (HR-pQCT) [Description]   \\
output & Measurement of trabecular and cortical bone microarchitecture  \\ [5pt]
\hdashline
input & [Endpoint/Outcome Measure] Bone Marrow Edema (BME) [Modality] X-Ray, MRI [Description]   \\
output & Detection of pathogenic changes in the bone marrow of the femoral head  \\ [5pt]
\hline
\end{tabular}
\end{figure*}

\newpage
\begin{figure*}[ht!]
\begin{tabular}[ht!]{|c|p{0.9\linewidth}|}
\hline
\multicolumn{2}{|c|}{\textbf{\textit{cluster\_tables : 3}}} \\
\hline
\multicolumn{2}{|c|}{\textbf{Task 1} (25 examples)} \\
\hline
input & [COOKIE name] CATEGORY\_INFO [COOKIE Description]   \\
output & Stores the category info on the page, that allows to display pages more quickly.  \\ [5pt]
\hdashline
input & [COOKIE name] FRONTEND [COOKIE Description]   \\
output & You sesssion ID on the server.  \\ [5pt]
\hdashline
input & [COOKIE name] CART [COOKIE Description]   \\
output & The association with your shopping cart.  \\ [5pt]
\hline
\multicolumn{2}{|c|}{\textbf{Task 2} (25 examples)} \\
\hline
input & [COOKIE name] WISHLIST\_CNT [COOKIE Description]   \\
output & The number of items in your Wishlist.  \\ [5pt]
\hdashline
input & [COOKIE name] NO\_CACHE [COOKIE Description]   \\
output & Indicates whether it is allowed to use cache.  \\ [5pt]
\hdashline
input & [COOKIE name] GUEST-VIEW [COOKIE Description]   \\
output & Allows guests to edit their orders.  \\ [5pt]
\hline
\multicolumn{2}{|c|}{\textbf{Task 3} (25 examples)} \\
\hline
input & [COOKIE name] CUSTOMER\_AUTH [COOKIE Description]   \\
output & An indicator if you are currently logged into the store.  \\ [5pt]
\hdashline
input & [COOKIE name] CUSTOMER [COOKIE Description]   \\
output & An encrypted version of your customer id with the store.  \\ [5pt]
\hdashline
input & [COOKIE name] STORE [COOKIE Description]   \\
output & The store view or language you have selected.  \\ [5pt]
\hline
\multicolumn{2}{|c|}{\textbf{Task 4} (25 examples)} \\
\hline
input & [COOKIE name] NO\_CACHE [COOKIE Description]   \\
output & Indicates whether it is allowed to use cache.  \\ [5pt]
\hdashline
input & [COOKIE name] LAST\_CATEGORY [COOKIE Description]   \\
output & The last category you visited.  \\ [5pt]
\hdashline
input & [COOKIE name] POLL [COOKIE Description]   \\
output & The ID of any polls you have recently voted in.  \\ [5pt]
\hline
\end{tabular}
\end{figure*}

\newpage
\begin{figure*}[ht!]
\begin{tabular}[ht!]{|c|p{0.9\linewidth}|}
\hline
\multicolumn{2}{|c|}{\textbf{\textit{nlp\_train}}} \\
\hline
\multicolumn{2}{|c|}{\textbf{numer\_sense} (100 examples)} \\
\hline
input & All scorpions have an additional [MASK] segments after the initial seven, ending in a sharp sting.  \\
output & five  \\ [5pt]
\hdashline
input & Heart failure affects about [MASK] million people in the United States.  \\
output & five  \\ [5pt]
\hdashline
input & Ribosomes have [MASK] subunits - small and large.  \\
output & two  \\ [5pt]
\hline
\multicolumn{2}{|c|}{\textbf{spider} (100 examples)} \\
\hline
input & What are the names of the climbers, ordered by points descending?  \\
output & SELECT Name FROM climber ORDER BY Points DESC  \\ [5pt]
\hdashline
input & Find the first names and offices of all instructors who have taught some course and also find the course description.  \\
output & SELECT T2.emp\_fname ,  T4.prof\_office ,  T3.crs\_description FROM CLASS AS T1 JOIN employee AS T2 ON T1.prof\_num  =  T2.emp\_num JOIN course AS T3 ON T1.crs\_code  =  T3.crs\_code JOIN professor AS T4 ON T2.emp\_num  =  T4.emp\_num  \\ [5pt]
\hdashline
input & What is the county that produces the most wines scoring higher than 90?  \\
output & SELECT T1.County FROM APPELLATIONS AS T1 JOIN WINE AS T2 ON T1.Appelation  =  T2.Appelation WHERE T2.Score  \textgreater{}  90 GROUP BY T1.County ORDER BY count(*) DESC LIMIT 1  \\ [5pt]
\hline
\multicolumn{2}{|c|}{\textbf{yahoo\_answers\_topics} (100 examples)} \\
\hline
input & question\_title: man date women but has serious secret interest exclusively in men who are women from waist up? [SEP] question\_content: and who wear make-up etc - is he really interested in men, and too afraid to come out of the closet or what? [SEP \textit{{... (Truncated)}} \\
output & Society \& Culture  \\ [5pt]
\hdashline
input & question\_title: bungee jumping site in victoria??? [SEP] question\_content: i am trying to find a site for bungee jumping around melbourne. i went thru the internet but couldnt find much. can anyone give me some info pls coz i ve been dreaming for t \textit{{... (Truncated)}} \\
output & Sports  \\ [5pt]
\hdashline
input & question\_title: celebs criminal conviction? [SEP] question\_content: can anybody suggesting some famous celebs or successful persons who's got criminal conviction? [SEP] best\_answer: Lots of celebrity activists have had criminal convictions, usuall \textit{{... (Truncated)}} \\
output & Politics \& Government  \\ [5pt]
\hline
\multicolumn{2}{|c|}{\textbf{piqa} (100 examples)} \\
\hline
input & goal: Preserve expensive lipstick. [SEP] solution 1Keep in clothes drawer. [SEP] solution 2Keep in fridge.  \\
output & 1  \\ [5pt]
\hdashline
input & goal: How to wash a dog. [SEP] solution 1Wet the dog with warm water, apply shampoo, lather and massage into fur, no need to rinse out all shampoo. Repeat process with conditioner if desired. [SEP] solution 2Wet the dog with warm water, apply shampoo \textit{{... (Truncated)}} \\
output & 1  \\ [5pt]
\hdashline
input & goal: To add a light inside a lamp. [SEP] solution 1Get wire with a plug, and chain, and feed the chain on. Then put on a washer -this should be decently big, and this is how the shade part will be attached. Then tape the wire to the socket, and scre \textit{{... (Truncated)}} \\
output & 1  \\ [5pt]
\hline
\end{tabular}
\end{figure*}

\newpage
\begin{figure*}[ht!]
\begin{tabular}[ht!]{|c|p{0.9\linewidth}|}
\hline
\multicolumn{2}{|c|}{\textbf{\textit{nlp\_test}}} \\
\hline
\multicolumn{2}{|c|}{\textbf{ag\_news} (100 examples)} \\
\hline
input & Delegation Is Delayed Before Reaching Najaf AGHDAD, Iraq, Aug. 17  A delegation of Iraqis was delayed for security reasons today but still intended to visit Najaf to try to convince a rebellious Shiite cleric and his militia to evacuate a shrine in t \textit{{... (Truncated)}} \\
output & World  \\ [5pt]
\hdashline
input & Restive Maldives eases curfew after rounding up dissidents (AFP) AFP - A curfew in the capital of the Maldives was eased but parliament sessions were put off indefinitely and emergency rule continued following last week's riots, officials and residen \textit{{... (Truncated)}} \\
output & World  \\ [5pt]
\hdashline
input & Another Major Non-Factor Another major, another disappointment for Tiger Woods, the No. 1 ranked player in the world who has not won a major championship since his triumph at the 2002 U.S. Open.  \\
output & Sports  \\ [5pt]
\hline
\multicolumn{2}{|c|}{\textbf{amazon\_polarity} (100 examples)} \\
\hline
input & title: Prompt shipment [SEP] content: I still haven't had time to watch the video to comment about the quality, but it was shipped promptly and seems to be in good order.  \\
output & positive  \\ [5pt]
\hdashline
input & title: Hey, we gotta talk [SEP] content: well, i gotta say this is one of her better albums. I'm real is da bomb and so is the I'm real (murder remix) she and ja rule sound SOOOOOO fine together. Love dont' cost a thing is hott too but Play is almost \textit{{... (Truncated)}} \\
output & positive  \\ [5pt]
\hdashline
input & title: absolute lemon [SEP] content: I probably have as much experience with 11x17 capable color printers as anyone in the world and I've got to say this is easily the most difficult and unsatisfactory printer I have ever dealt with. HP's last 11x17  \textit{{... (Truncated)}} \\
output & negative  \\ [5pt]
\hline
\multicolumn{2}{|c|}{\textbf{commonsense\_qa} (100 examples)} \\
\hline
input & What is the main purpose of farmers?  \\
output & supply food  \\ [5pt]
\hdashline
input & When drinking booze what can you do to stay busy?  \\
output & examine thing  \\ [5pt]
\hdashline
input & If you are prone to postpone work what will you have to do in order to finish on time?  \\
output & hasten  \\ [5pt]
\hline
\multicolumn{2}{|c|}{\textbf{dbpedia\_14} (100 examples)} \\
\hline
input & Edward Albert Eddie Nicholls (born 10 December 1947 in British Guiana) is an international Test and One Day International cricket umpire. Nicholls officiated in 17 Tests from 1997 to 2001 and 46 ODIs from 1995 to 2005. Nicholls is one of four West In \textit{{... (Truncated)}} \\
output & Athlete  \\ [5pt]
\hdashline
input & Cryptolepis is a plant genus in the family Apocynaceae. It includes some 42 species.  \\
output & Plant  \\ [5pt]
\hdashline
input & Mike Enneking is a retired American soccer player who played professionally in the Western Soccer Alliance and Major Indoor Soccer League. He is the Vice President of Crossfire Premier Soccer Club.Enneking graduated from Bellarmine Preparatory School \textit{{... (Truncated)}} \\
output & Athlete  \\ [5pt]
\hline
\end{tabular}
\end{figure*}

\newpage
\begin{figure*}[ht!]
\begin{tabular}[ht!]{|c|p{0.9\linewidth}|}
\hline
\multicolumn{2}{|c|}{\textbf{\textit{nlp\_test}}} \\
\hline
\multicolumn{2}{|c|}{\textbf{ethos-race} (87 examples)} \\
\hline
input & In my dream I wished you have died! Don't call yourself man, while talking like queen.  \\
output & false  \\ [5pt]
\hdashline
input & Is that your nose or is that Thor's Hammer?  \\
output & false  \\ [5pt]
\hdashline
input & All girls like you should have a pimp to guide them to successful prostitution  \\
output & false  \\ [5pt]
\hline
\multicolumn{2}{|c|}{\textbf{financial\_phrasebank} (100 examples)} \\
\hline
input & Profit before taxes was EUR 5.4 mn , up from EUR 3.6 mn a year earlier .  \\
output & positive  \\ [5pt]
\hdashline
input & The order was valued at USD12 .2 m.  \\
output & neutral  \\ [5pt]
\hdashline
input & The company expects net sales to significantly increase from 2009 .  \\
output & positive  \\ [5pt]
\hline
\multicolumn{2}{|c|}{\textbf{qasc} (100 examples)} \\
\hline
input & what is tourette syndrome?  \\
output & trait  \\ [5pt]
\hdashline
input & Animals that are \_ provide little if any care to their young.  \\
output & cold blooded  \\ [5pt]
\hdashline
input & What can be used for transportation?  \\
output & trailers and boats  \\ [5pt]
\hline
\multicolumn{2}{|c|}{\textbf{sciq} (100 examples)} \\
\hline
input &  All alkaline Earth metals have similar properties because they all have two valence electrons. They readily give up their two valence electrons to achieve a full outer energy level, which is the most stable arrangement of electrons. As a result, the \textit{{... (Truncated)}} \\
output & valence electrons  \\ [5pt]
\hdashline
input &  Exposure gives an indication of the amount of radiation that travels through the air. Two factors influence the amount of exposure a person may receive – time and intensity. Acute exposure indicates a large amount of radiation received over a short  \textit{{... (Truncated)}} \\
output & chronic exposure  \\ [5pt]
\hdashline
input &  Ventricular Systole Ventricular systole (see Figure 19.27) follows the depolarization of the ventricles and is represented by the QRS complex in the ECG. It may be conveniently divided into two phases, lasting a total of 270 ms. At the end of atrial \textit{{... (Truncated)}} \\
output & pulmonary and aortic semilunar  \\ [5pt]
\hline
\end{tabular}
\end{figure*}

\newpage
\begin{figure*}[ht!]
\begin{tabular}[ht!]{|c|p{0.9\linewidth}|}
\hline
\multicolumn{2}{|c|}{\textbf{\textit{nlp\_test}}} \\
\hline
\multicolumn{2}{|c|}{\textbf{tweet\_eval-stance\_atheism} (52 examples)} \\
\hline
input & The worst day of my life so far is here, setting my Nan to rest. Even as a physicist, times like these make you wonder. \#SemST  \\
output & none  \\ [5pt]
\hdashline
input & I will dwell in a peaceful habitation, in secure dwellings, and in quiet resting places -Isa. 32:18 \#SemST  \\
output & against  \\ [5pt]
\hdashline
input & @user sweet! Congratulations to a rational decision. \#SemST  \\
output & none  \\ [5pt]
\hline
\multicolumn{2}{|c|}{\textbf{yelp\_polarity} (100 examples)} \\
\hline
input & Very disappointed in this salon. Set an appt 4 days ahead of time. Area were I for my set put on was dirty from a past client. The mail tech did not talk, I felt rushed through my appt which resulted in me leaving unhappy. I won't be returning.  \\
output & negative  \\ [5pt]
\hdashline
input & Our flight arrived to Vegas earlier than excepted, so we expected our room not to be ready. When we arrived at the hotel on May 19th, the front desk girl offered us a room that was ready on the 28th floor that wasn't facing the Bellagio fountain. I b \textit{{... (Truncated)}} \\
output & positive  \\ [5pt]
\hdashline
input & My poor children who live out of state, have no idea how cheap and ugly the flowers I just received from Carmel Florist are. They do not resemble the online photo at all. I actually laughed at the gentleman who delivered them to my door. They spent \ \textit{{... (Truncated)}} \\
output & negative  \\ [5pt]
\hline
\end{tabular}
\end{figure*}

\end{document}